\documentclass{article}

\usepackage[final,nonatbib]{neurips_data_2024}
\usepackage{paralist}
\usepackage{subcaption}

\usepackage[utf8]{inputenc} %
\usepackage[T1]{fontenc}    %
\usepackage{url}            %
\usepackage{booktabs}       %
\usepackage{amsfonts}       %
\usepackage{nicefrac}       %
\usepackage{microtype}      %
\usepackage{wrapfig}        %
\usepackage[numbers,compress]{natbib}

\usepackage{enumitem}
\usepackage{makecell}

\usepackage{graphicx} %
\usepackage[table,xcdraw]{xcolor}
\usepackage{tabularx, makecell}
\usepackage{listings}
\usepackage{mdframed}

\usepackage{tikz}
\usepackage{hyperref}       %
\usepackage{cleveref}

\definecolor{codegreen}{rgb}{0,0.6,0}
\definecolor{codegray}{rgb}{0.5,0.5,0.5}
\definecolor{codepurple}{rgb}{0.58,0,0.82}
\definecolor{backcolour}{rgb}{0.95,0.95,0.92}

\definecolor{delim}{RGB}{20,105,176}
\definecolor{numb}{RGB}{106, 109, 32}
\definecolor{string}{rgb}{0.64,0.08,0.08}

\def\tikzmark#1{\tikz[remember picture,overlay]\node[yshift=2pt](#1){};}

\renewcommand{\footnotesize}{\fontsize{8pt}{11pt}\selectfont}
\newcommand*{\yellowemph}[1]{%
\tikz[baseline]\node[rectangle, fill=yellow!20, rounded corners, inner sep=0.3mm,anchor=base]{#1};%
}
\newcommand*{\orangeemph}[1]{%
\tikz[baseline]\node[rectangle, fill=orange!20, rounded corners, inner sep=0.3mm,anchor=base]{#1};%
}

\lstdefinelanguage{json}{
    backgroundcolor=\color{backcolour},   
    commentstyle=\color{codegreen},
    keywordstyle=\color{magenta},
    numberstyle=\tiny,
    stringstyle=\color{codepurple},
    basicstyle=\ttfamily\footnotesize,
    breakatwhitespace=false,         
    breaklines=true,                 
    captionpos=b,                    
    keepspaces=true,                 
    numbers=left,                    
    numbersep=5pt,                  
    showspaces=false,                
    showstringspaces=false,
    showtabs=false,                  
    tabsize=2,
    numbers=left,
    numberstyle=\tiny,
    rulecolor=\color{black},
    showspaces=false,
    showtabs=false,
    breaklines=true,
    breakatwhitespace=true,
    basicstyle=\ttfamily\footnotesize,
    upquote=true,
    morestring=[b]",
    stringstyle=\color{string},
    escapeinside={*@}{@*},
    literate=
     *{0}{{{\color{numb}0}}}{1}
      {1}{{{\color{numb}1}}}{1}
      {2}{{{\color{numb}2}}}{1}
      {3}{{{\color{numb}3}}}{1}
      {4}{{{\color{numb}4}}}{1}
      {5}{{{\color{numb}5}}}{1}
      {6}{{{\color{numb}6}}}{1}
      {7}{{{\color{numb}7}}}{1}
      {8}{{{\color{numb}8}}}{1}
      {9}{{{\color{numb}9}}}{1}
      {\{}{{{\color{delim}{\{}}}}{1}
      {\}}{{{\color{delim}{\}}}}}{1}
      {[}{{{\color{delim}{[}}}}{1}
      {]}{{{\color{delim}{]}}}}{1},
}

\title{Croissant: A Metadata Format for ML-Ready Datasets} %

\author{%
Mubashara~Akhtar\textsuperscript{1}\textsuperscript{*},
Omar~Benjelloun\textsuperscript{2}\textsuperscript{*},
Costanza~Conforti\textsuperscript{2}\textsuperscript{*},
Luca~Foschini\textsuperscript{3}\textsuperscript{*},\\
\textbf{%
Pieter~Gijsbers\textsuperscript{4},
Joan~Giner-Miguelez\textsuperscript{5,22}\textsuperscript{*},
Sujata~Goswami\textsuperscript{6},
Nitisha~Jain\textsuperscript{1}\textsuperscript{*},
}\\
\textbf{%
Michalis~Karamousadakis\textsuperscript{7},
Satyapriya~Krishna\textsuperscript{8},
Michael~Kuchnik\textsuperscript{9}\textsuperscript{*},
Sylvain~Lesage\textsuperscript{10}\textsuperscript{*},
}\\
\textbf{%
Quentin~Lhoest\textsuperscript{10}\textsuperscript{*},
Pierre~Marcenac\textsuperscript{2}\textsuperscript{*},
Manil~Maskey\textsuperscript{11},
Peter~Mattson\textsuperscript{2},
Luis~Oala\textsuperscript{12}\textsuperscript{*},
}\\
\textbf{%
Hamidah~Oderinwale\textsuperscript{13},
Pierre~Ruyssen\textsuperscript{2}\textsuperscript{*},
Tim~Santos\textsuperscript{14},
Rajat~Shinde\textsuperscript{15}\textsuperscript{*},
Elena~Simperl\textsuperscript{1,16}\textsuperscript{*},
}\\
\textbf{%
Arjun~Suresh\textsuperscript{17},
Goeffry~Thomas\textsuperscript{2,18}\textsuperscript{*},
Slava~Tykhonov\textsuperscript{19}\textsuperscript{*},
Joaquin~Vanschoren\textsuperscript{4}\textsuperscript{*},
}\\
\textbf{%
Susheel~Varma\textsuperscript{3},
Jos~van~der~Velde\textsuperscript{4}\textsuperscript{*},
Steffen~Vogler\textsuperscript{20},
Carole-Jean~Wu\textsuperscript{9},
Luyao~Zhang\textsuperscript{21}
}\\
Authors in alphabetical order
}

\begin{document}

\maketitle
\textsuperscript{*}Core contributors
\textsuperscript{1}King’s College London,
\textsuperscript{2}Google,
\textsuperscript{3}Sage Bionetworks,
\textsuperscript{4}Eindhoven University of Technology,
\textsuperscript{5}Universitat Oberta de Catalunya,
\textsuperscript{6}Oak Ridge National Laboratory,
\textsuperscript{7}Plaixus Ltd,
\textsuperscript{8}Harvard University,
\textsuperscript{9}Meta,
\textsuperscript{10}Hugging Face,
\textsuperscript{11}NASA,
\textsuperscript{12}Dotphoton,
\textsuperscript{13}McGill University,
\textsuperscript{14}Graphcore,
\textsuperscript{15}NASA IMPACT \& UAH,
\textsuperscript{16}Open Data Institute,
\textsuperscript{17}GATE Overflow, India,
\textsuperscript{18}Kaggle,
\textsuperscript{19}DANS-KNAW,
\textsuperscript{20}Bayer,
\textsuperscript{21}Duke Kunshan University,
\textsuperscript{22}Barcelona Supercomputing Center (BSC) 

\begin{abstract}%

Data is a critical resource for machine learning (ML), yet working with data remains a key friction point. This paper introduces Croissant,  a metadata format for datasets that creates a shared representation across ML tools, frameworks, and platforms.
Croissant makes datasets more discoverable, portable, and interoperable, thereby addressing significant challenges in ML data management. Croissant is already supported by several popular dataset repositories, spanning hundreds of thousands of datasets, enabling easy loading into the most commonly-used ML frameworks, regardless of where the data is stored. Our initial evaluation by human raters shows that Croissant metadata is readable, understandable, complete, yet concise.%
\end{abstract}%
\section{Introduction}%
\label{sec:intro}
Recent machine learning (ML) advances highlight the critical role of data management in achieving technological breakthroughs. Yet, working with data remains time-consuming and painful due to a wide variety of data formats, the lack of interoperability between tools, and the difficulty of discovering and combining datasets \cite{kuchnik2022plumber,2024dmlr}.  Data's prominent role in ML also leads to questions about its responsible use for training and evaluating ML models in areas such as licensing, privacy, or fairness, among others~\cite{sambasivan2021everyone}. New approaches are needed to make datasets easier to work with, while also addressing concerns around their responsible use.
    
\begin{wrapfigure}{r}{0.6\textwidth}
    \vspace{-.4cm}
  \begin{center}
    \includegraphics[width=0.5\textwidth]{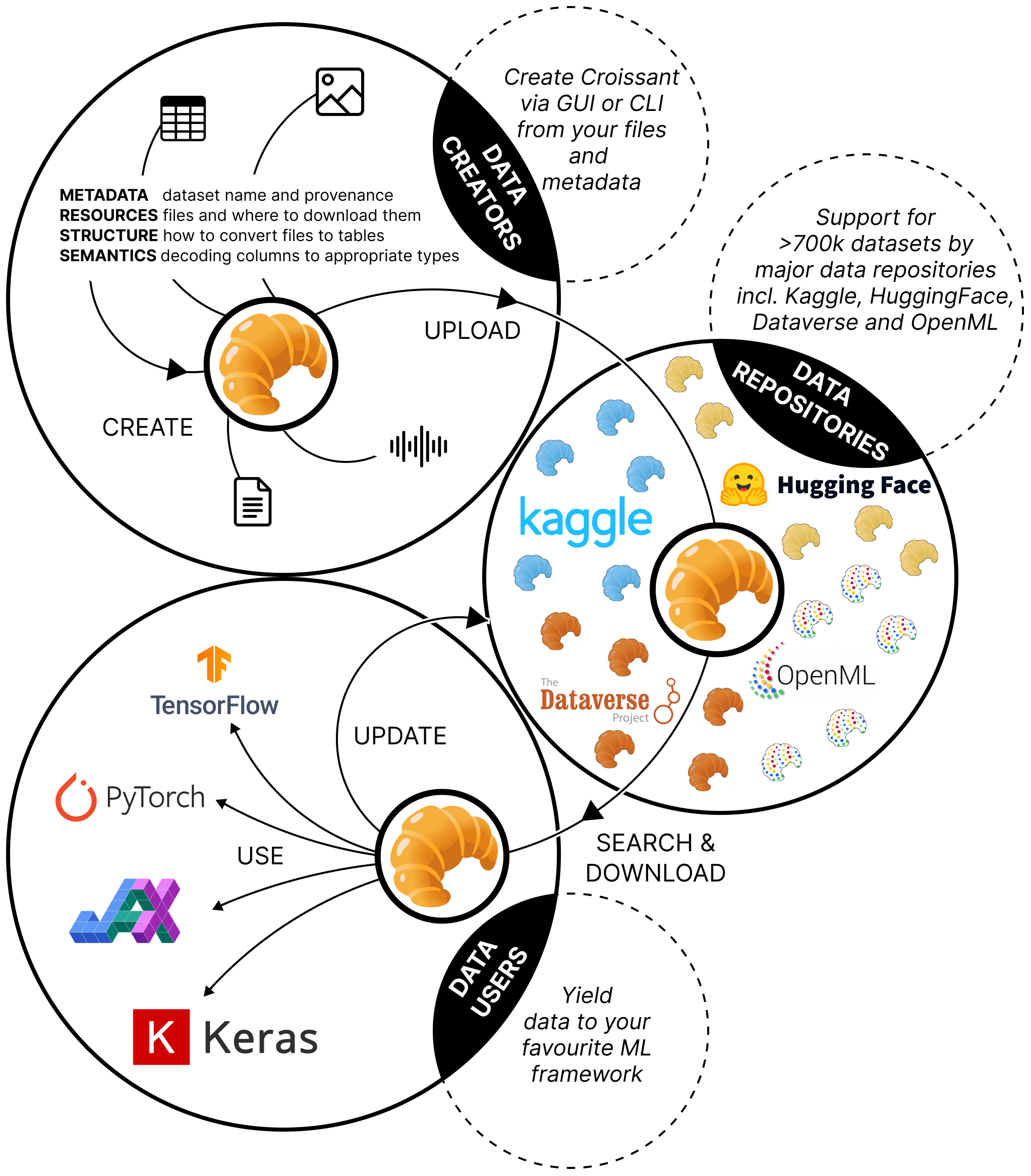}
  \end{center}
  \caption{The Croissant lifecycle and ecosystem.}
  \vspace{-1.5cm}
  \label{fig:abstract}
\end{wrapfigure}

This paper\footnote{A shorter preliminary introduction to Croissant was presented at the DEEM 2024 workshop \cite{10.1145/3650203.3663326}.} presents \textit{Croissant}, a metadata format designed to improve ML datasets' discoverability, portability, reproducibility, and interoperability.  Croissant makes datasets ``ML-ready'' by recording ML-specific metadata that enables them to be loaded directly into ML frameworks and tools (see Figure~\ref{fig:code} for sample code). Croissant describes datasets' attributes, the resources they contain, and their structure and semantics. This uniform description streamlines their usage and sharing within the ML community and between ML platforms and tools while fostering responsible ML practices. Figure~\ref{fig:abstract} gives an overview of the Croissant lifecycle and ecosystem.

Croissant can describe most types of data commonly used in ML workflows, such as images, text, audio, or tabular. While datasets come in a variety of data formats and layouts, Croissant exposes a unified ``view'' over these resources. It lets users add semantic descriptions and ML-specific information. The Croissant vocabulary~\cite{croissant_spec} does not require changing the underlying data representation, and can thus be easily added to existing datasets, and adopted by dataset repositories.

To assess Croissant's usability, we conducted a preliminary usability evaluation on metadata creation for language, vision, audio, and multi-modal datasets. Several practitioners annotated ten widely used ML datasets. We analyzed the consistency of their responses and collected their feedback on Croissant.

The remainder of the paper is structured as follows: in Section 2 we contextualize related work. In Section 3 we describe the Croissant format, its integrations, and the tools that support it. Section 4 comprises the user study and discusses its results and limitations.

\section{Related Work}%

While there have been many prior efforts in standardizing dataset metadata, they typically lack ML-specific support, do not work with existing ML tools, or lag behind the demands of dynamically evolving requirements, such as responsible ML.
We outline the state of the field below.

\textbf{Vocabularies for Dataset Documentation.}
Dataset documentation is indispensable for effective data management and serves as a foundational element for training and evaluating ML models~\cite{gebru2021datasheets}. Metadata descriptions of datasets enhance their discoverability, interoperability, and usability, which is critical for advancing research and data-driven applications. Ontologies and vocabularies are semantic web tools used to standardize dataset documentation.
While vocabularies comprise sets of terms and their meanings to describe data consistently, ontologies provide a structured framework to define and relate these concepts within a domain. Ontologies and vocabularies are evaluated for their coverage (i.e., do they represent all relevant concepts), accuracy (correctness of definitions and relationships), consistency (no logical contradictions), and usability (ease of use and integration). This is done through methods like competency questions, expert validation, and use-case testing~\cite{wilson2022conceptual}.

\textbf{Standards for Catalogs and Metadata.} With the increase of data availability online, various efforts have focused on making data both discoverable and user-friendly by supplementing datasets with comprehensive metadata. This metadata may include details about the dataset, such as authorship, format, and intended use, all structured consistently to support automated processing and retrieval.
Key efforts towards documentation have led to the creation of standards like the Data Catalog Vocabulary (DCAT)~\cite{DCAT} and the \texttt{Dataset} vocabulary in \texttt{schema.org}~\cite{schema_dataset}. DCAT facilitates interoperability among web-based data catalogs, enabling users to aggregate, classify, and filter datasets efficiently. \texttt{Schema.org}~\cite{guha2016schema} acts as a de facto standard for metadata, helping search engines discover and index published web content, including datasets, thus enhancing dataset accessibility and understandability. This versatility allows \texttt{schema.org} to describe a wide array of content types effectively.
Other frameworks, such as Data Packages~\cite{data-packages} and CSV on the Web~\cite{csvw} support methods for describing and exchanging tabular data.
The Global Alliance for Genomics and Health's Data Use Ontology (DUO)~\citep{lawson2021data} refines data usage terms with optional modifiers, improving clarity in genomic data sharing agreements. Efforts towards integration of FAIR principles (Findability, Accessibility, Interoperability, and Reusability)~\cite{wilkinson2016fair} in metadata vocabularies are also noteworthy.
Despite their utility for specific domains and formats, these standards do not entirely meet the specialized needs of data management within the ML domain. In this context, the compliance of ML-ready datasets with the FAIR principles is a primary need for improving discoverability, portability and
reproducibility in the ML ecosystem. The adoption of standard metadata description practices across the broader community further enhances the interoperability of ML datasets from diverse domains.

\textbf{Operationalizing Responsible ML through Data Work.}
Data-centric ML~\cite{2024dmlr,data-centric} is increasingly seen as critical to the development of trustworthy ML systems, including aspects such as fairness, accountability, transparency, data privacy and governance, safety, and robustness~\citep{Smuha2019}.
Seminal works, such as Datasheets for Datasets~\cite{gebru2021datasheets} and Data Statements~\cite{bender-friedman-2018-data}, have emphasized the importance of dataset documentation
to assess and increase the trustworthiness of ML systems.
Several related documentation efforts such as Data Cards~\citep{pushkarna2022data} and Data Nutrition Labels~\cite{holland2018dataset} have been inspired them. 
ML data repositories, such as Kaggle~\cite{kaggle_datasets}, OpenML~\cite{OpenML2013} and Hugging Face~\cite{huggingface_datasets}, have initiated their own metadata documentation efforts. Hugging Face, for example, provides Dataset Cards~\cite{hf_datasets_cards} that include summaries, fields, splits, potential social impacts, and biases inherent in the datasets.

\begin{figure}[t]
\begin{mdframed}[backgroundcolor=black!5,leftmargin=0.27cm,skipabove=0.3cm,hidealllines=true,%
  innerleftmargin=0.1cm,innerrightmargin=0.2cm,innertopmargin=-0.0cm,innerbottommargin=-0.10cm]
\begin{lstlisting}[language=Python]
# 1. Point to a local or remote Croissant JSON file
import mlcroissant as mlc
url = "https://huggingface.co/api/datasets/fashion_mnist/croissant"
# 2. Inspect metadata
print(mlc.Dataset(url).metadata.to_json())
# 3. Use Croissant dataset in your ML workload
import tensorflow_datasets as tfds
builder = tfds.core.dataset_builders.CroissantBuilder(
    jsonld=url, file_format="array_record")
builder.download_and_prepare()
# 4. Split for training/testing
train, test = builder.as_data_source(
    split=["default[:80%
\end{lstlisting}
\end{mdframed}
\vspace{-.3cm}
\caption{
Users can easily inspect datasets (e.g., Fashion MNIST~\cite{fashion-mnist}) and use them in data loaders with Croissant. See Supplementary material or visit \url{https://github.com/mlcommons/croissant} for more examples.
}
\label{fig:code}
\end{figure}

These approaches typically rely on data documentation written in natural language, without a standard machine-readable representation, which makes data documentation challenging for machines to read and process.
Croissant fills this gap by providing a standardized framework for data documentation that ensures semantic consistency and machine readability, thereby facilitating seamless integration with existing tools and frameworks used by the ML community.

\section{The Croissant Format}%

The Croissant format is a community-driven metadata vocabulary for describing datasets that builds on Schema.org \cite{guha2016schema}. Croissant is divided into four layers:
\begin{inparaenum}[\itshape(i)\upshape] \item The \emph{Dataset Metadata Layer}, containing relevant information such as name, description, and version. \item The \emph{Resource Layer} describes the source data used in the dataset. \item The \emph{Structure Layer}, describing and organizing the structure of the resources. \item The \emph{Semantic Layer}, which provides ML-specific data interpretation and semantics. \end{inparaenum} 
A more detailed description of the Croissant format can be found in the official specification \cite{croissant_spec}.
Documentation and code is available online\footnote{\href{https://docs.mlcommons.org/croissant/}{https://docs.mlcommons.org/croissant/}}.

In the remainder of this section, we illustrate each layer with examples from popular ML datasets. Afterwards, we briefly describe the Croissant Responsible AI extension, and then provide an overview of ML frameworks, tools, and repositories that currently support Croissant.

\subsection{The Dataset Metadata Layer}
\label{ssec:dataset_layer}

Croissant dataset descriptions, illustrated in Figure~\ref{fig:dataset-resources}, are based on \path{schema.org/Dataset}, a widely adopted vocabulary for datasets on the Web~\cite{schema_dataset}, hence ensuring interoperability with existing standards and tools. Croissant specifies constraints on which \texttt{schema.org} properties are required, recommended and optional, and adds additional properties, e.g., to represent snapshots, live datasets, and citation information.

\subsection{The Resources Layer}
This layer represents the data resources (e.g., files) of the dataset. \texttt{Schema.org} properties are insufficient to adequately describe dataset contents with complex layouts, which are common for ML datasets. This layer provides two primitive classes to address this limitation and describe dataset resources: \texttt{FileObject} to describe individual files and \texttt{FileSet} to describe sets of files.

Figure \ref{fig:dataset-resources} shows an excerpt of the Croissant definition of the PASS dataset \cite{asano21pass}, where declarations of object names are highlighted in \yellowemph{yellow}, with references in \orangeemph{orange}. This distribution includes two \texttt{FileObject}s: a CSV file containing metadata about the dataset (line 13) and an archive file containing images (line 20). Moreover, \texttt{FileSet} (in line 27) is used to refer to a collection of images, videos, or text files that contain the (unlabeled) data used for training and inference. Since there can be numerous files, \texttt{FileSets} are specified with inclusion/exclusion filters (e.g., a pattern matching all files that should be included) as shown on line 30. 

\begin{figure}[t]

   \begin{minipage}[t]{0.48\textwidth}
\centering

\begin{lstlisting}[language=json, basicstyle=\scriptsize]
{ 
  "@type": "sc:Dataset",
  "name": "PASS",
  "dct:conformsTo": "http://mlcommons.org/croissant/1.0",
  "description": 
   . "PASS is a large-scale image dataset...",
  "citeAs": "@Article{asano21pass, ...",
  "license": "cc-by-4.0",
  "url": "https://www.robots.ox.ac.uk/.../pass/"
  
  "distribution": [
  {
    "@id": *@\yellowemph{"metadata"}@*,
    "@type": "cr:FileObject",
    "contentUrl": "https://zenodo.org/661...",
    "sha256": "0b033707ea49365a5ffdd1461...",
    "encodingFormat": "text/csv"
  },
  { 
    "@id": *@\yellowemph{"pass0"}@*,
    "@type": "cr:FileObject",
    "contentUrl": "https://zenodo.org/661...",
    "sha256": "0be3a104d6257d83296460b...",
    "encodingFormat": "application/x-tar"
  },
  { 
    "@id": *@\yellowemph{"image-files"}@*,
    "@type": "cr:FileSet",
    "containedIn": { "@id":*@\orangeemph{"pass0"}@* }
    "includes": "*.jpg",
    "encodingFormat": "image/jpeg"
  }],
}





 
 
 \end{lstlisting}
\caption{Dataset metadata and resources for the PASS dataset. }
\label{fig:dataset-resources}
\end{minipage}\qquad
   \begin{minipage}[t]{0.48\textwidth}
\begin{lstlisting}[language=json, basicstyle=\scriptsize]
{ "@id": *@\yellowemph{"images"}@*,
  "@type": "cr:RecordSet",
  "key": "images/hash", *@\tikzmark{A}@*
  "field": [
    { "@id": *@\yellowemph{"images/image\_content"}@*,
      "@type": "cr:Field",
      "dataType": "sc:ImageObject",
      "source": {
        "fileSet":{"@id": *@\orangeemph{"image-files"}@*},
        "extract":{"fileProperty":"content"}
      }
    },
    {
      "@id": *@\yellowemph{"images/hash"}@*, *@\tikzmark{B}@*
      "@type": "cr:Field",
      "dataType": "sc:Text",
      "source": {
        "fileSet": {"@id": *@\orangeemph{"image-files"}@*},
        "extract": {"fileProperty": "filename"},
        "transform": {"regex": "([^\\/]*)\\.jpg"}
      },
      "references": {
        "fileObject": {"@id": *@\orangeemph{"metadata"}@*},
        "column": "hash"
      }
    },    
    { "@id": *@\yellowemph{"images/coordinates"}@*,
      "@type": "cr:Field",
      "dataType": "sc:GeoCoordinates",
      "subField": [
       {  "@id": *@\yellowemph{"images/coordinates/latitude"}@*,
          "@type": "cr:Field",
          "source": {
            "fileObject": {"@id": *@\orangeemph{"metadata"}@*},
            "column": "latitude"}
       },
       { "@id": *@\yellowemph{"images/coordinates/longitude"}@*,
        "@type": "cr:Field",
        "source": {
            "fileObject": {"@id": *@\orangeemph{"metadata"}@*},
            "column": "longitude"}
       }]
    }]
}  
\end{lstlisting}
\caption{A \texttt{RecordSet} that joins images and structured metadata from the PASS dataset.}
\label{fig:dataset-recordsets}
   \end{minipage}
\end{figure}

\subsection{The Structure Layer}

While \texttt{FileObject} and \texttt{FileSet} describe a dataset's resources, they lack information on how the content of the resources is organized. This is addressed with \texttt{RecordSet}, which allows loading data of various formats into a standard representation, including structured (CSV and JSON) and unstructured (text, audio, and video) data.
Handling all data formatting information in one layer abstracts away format heterogeneity, addressing a key challenge in processing and loading ML data.

\texttt{RecordSet} provides a common structure description for records that may contain multiple fields, which can be used across different modalities. As an example, Figure~\ref{fig:dataset-recordsets} shows a \texttt{RecordSet} combining images from PASS with additional features from a metadata CSV file. Each \texttt{Field} in the \texttt{RecordSet} defines the source of its data, which may refer to the contents of elements in a \texttt{FileSet}. For instance, the \texttt{Field images/image\_content} in line 9 refers to the \texttt{image-files} \texttt{FileSet} and also points to the specific property to extract in line 10.

 \texttt{Field}s can be nested, as we can see in the \texttt{images/coordinates} field, which contains two subfields: \texttt{images/coordinates/latitude} and \texttt{images/coordinates/longitude}. Croissant supports nesting entire \texttt{RecordSet}s, e.g., to add annotations (e.g. object bounding boxes) to images, where each image may correspond to multiple structured annotations. See Croissant's COCO~\cite{coco} definition\footnote{\url{https://github.com/mlcommons/croissant/blob/main/datasets/1.0/coco2014/metadata.json}} for a representative example. \texttt{RecordSet} also supports joining heterogeneous data and data manipulation methods, like JSON Path and regular expressions, for flexible data extraction and transformation.

\subsection{The Semantic Layer}

The semantic layer introduces a number of useful features in the context of ML data. These are implemented using the primitives defined in the previous sections, generally as new classes or properties defined in the Croissant namespace. Semantic typing is used to describe important aspects of ML practice, such as the dataset splits (train, test, validation) as well as dataset labels. Additionally, semantic typing is used to describe commonly used data types, such as bounding boxes, categorical data, or segmentation masks. %
As an example, in Figure \ref{fig:dataset-recordsets}, the structured \texttt{Field} \texttt{images/coordinates} has the dataType \texttt{GeoCoordinates}\footnote{\url{http://schema.org/GeoCoordinates}} from schema.org. The subFields \texttt{images/coordinates/latitude} and \texttt{images/coordinates/longitude} are implicitly mapped to the latitude and longitude properties associated with that class, because their names \mbox{match by suffix.}

\subsection{The Croissant-RAI Extension}
\label{ssec:croissant_rai}

Croissant-RAI~\citep{croissant_rai} is an extension of the Croissant format that builds on existing responsible AI (RAI) dataset documentation approaches, such as Data Cards~\citep{pushkarna2022data} and Datasheets for Datasets~\citep{gebru2021datasheets}, making it easier to publish, discover, and reuse RAI metadata. The extension was developed around RAI use cases such as documenting the data life cycle, data labeling and participatory processes, information for AI safety, fairness assessments, and regulatory compliance.
It was developed through a multi-step, iterative vocabulary engineering process. Based on the target use cases, a list of properties was defined through evaluation of related dataset documentation vocabularies and the Croissant vocabulary with an aim to detect overlaps and gaps. The resulting properties were evaluated by annotating example datasets to verify their usability and usefulness. For more details, see~\citep{Jain24}.

\subsection{Croissant Tools and Integrations}
\label{ssec:tools_integration}

In parallel with the definition of the Croissant format, we have pursued a number of integrations, with the goals of 1) making Croissant immediately useful to users, and 2) grounding Croissant in the requirements of real-world datasets and tools. Figure ~\ref{fig:abstract} gives an overview of the Croissant ecosystem. %

\paragraph{Data Repositories.} Croissant has been integrated into three major dataset repositories: Hugging Face Datasets, Kaggle Datasets, and OpenML, which together describe over 400,000 datasets in the Croissant format. This integration has succeeded with minimal effort because Croissant is an extension of the widely adopted \texttt{Schema.org/Dataset} vocabulary and does not require changing the existing data layout. Supporting Croissant involved adding additional fields to existing metadata. Furthermore, most repositories offer normalized data representations (Hugging Face and OpenML convert most datasets to Parquet) and their own data types (such as relational schemas for tabular data). Consequently, the conversion to Croissant primarily focuses on managing these data formats and specifying associated data types as \texttt{RecordSet} definitions.

In addition to the support from individual data repositories, Croissant is also supported by Google Dataset Search~\cite{dataset_search}. When a user searches for a query that returns Croissant datasets, a special filter allows them to restrict the results to only Croissant datasets. This functionality allows users to effectively search for Croissant datasets across data repositories and the entire web.

\paragraph{ML Frameworks.}

Croissant’s reference implementation is a standalone Python library%
that supports the validation of Croissant dataset descriptions, their programmatic creation and manipulation, and serialization into JSON-LD. To consume data, the library provides an iterator abstraction that interoperates with existing data loaders. 
The TensorFlow Datasets~\cite{TFDS} library provides a dataset builder\footnote{\url{https://www.tensorflow.org/datasets/format\_specific\_dataset\_builders\#croissantbuilder}} that prepares the dataset on disk in a format compatible with JAX, TensorFlow and PyTorch loaders. Alternatively, frameworks such as PyTorch DataPipes~\cite{pytorch_datapipes} interface with the Croissant library by wrapping the iterator directly. We anticipate that additional optimization opportunities will arise with more varied and larger datasets, perhaps requiring distributed execution as well as more advanced operator scheduling.

\textbf{Croissant Editor.}
Croissant is primarily a machine readable format (in JSON-LD), so users may find it hard to create dataset descriptions by hand. We developed the Croissant Editor\footnote{\url{https://huggingface.co/spaces/MLCommons/croissant-editor}}, 
(also on GitHub\footnote{\url{https://github.com/mlcommons/croissant/tree/main/editor}}), 
a tool that lets users visually create and modify Croissant datasets. The Croissant Editor provides form-based editing and validation of Croissant metadata, and bootstraps the definition of resources and \texttt{RecordSet}s by inferring them from the data uploaded by the user. The editor integrates  the Croissant Responsible AI extension, and guides users in describing RAI aspects of their datasets. 

\subsection{The Croissant Working Group}

We designed the Croissant format in an open and participatory way. The MLCommons Croissant Working Group (WG)\footnote{See the MLCommons website for further details: \url{https://mlcommons.org/working-groups/data/croissant/}} consists of diverse stakeholders and domain experts from academia, industry, research organizations, and collaborative networks such as the AI for Public Good network.
Use cases were discussed and presented to WG members (including domain experts) as they were developed, ensuring that diverse views and priorities were covered. The schema is designed to be modular and extensible, allowing for domain-specific attributes and concerns to be integrated into the Core Croissant format.
We continuously collect feedback from working group members and users and are committed to incorporating this feedback in future versions of Croissant. Additionally, Croissant is based on \texttt{schema.org}, a well-established vocabulary. 
\section{Croissant Evaluation: A User Study with ML Practitioners}%
This section describes the user study we conducted to evaluate the Croissant metadata format. We asked machine learning practitioners to annotate a variety of datasets commonly used in the ML community. Human annotators authored a subset of the Croissant and Croissant-RAI attributes and assessed them based on criteria commonly used in vocabulary evaluation~\citep{Gomez-Perez01}. 

\subsection{The User Study Process}
\label{ssec:user_study}

\paragraph{Recruitment of Annotators and Annotation Process.}

We recruited nine volunteers from the Croissant development community who were all proficient in English with backgrounds in vocabulary and ontology engineering, dataset documentation, ML benchmarking, and responsible AI. We collected demographic information from all annotators, which we published in the user study report~\citep{croissant_working_group_2024_13350974}.
For each one of the ten datasets, we collected metadata definitions from three annotators, resulting in thirty annotations. Each human annotator assessed approximately three datasets on average, with three annotating one dataset and one person annotating six datasets.\footnote{We publish the user study specifications and collected data \citep{croissant_working_group_2024_13350974}.}

The instructions for the annotators were comprised of: $(i)$ a short introduction to the Croissant metadata format; $(ii)$ the purpose of the user study; $(iii)$ the definitions of the requested Croissant and Croissant-RAI attributes; $(iv)$ links to the format specifications, and $(v)$ a link to each dataset in the Hugging Face repository.
Prior to starting the user study, we obtained ethical clearance and informed annotators about the data being collected and its purpose.
For each dataset, annotators filled out a provided JSON template with the sixteen attributes to complete.
Afterwards, annotators answered questions about their level of understanding of the datasets (see~\Cref{tab:evaluation_questions}), 
and indicated their confidence in the annotations they provided on a Likert scale~\citep{likert1932technique} between $1$ and $5$. We followed previous research~\citep{troiano-etal-2021-emotion} suggesting that confidence ratings can serve as a tool to understand potential annotation inconsistencies. 
The user study began in April $2024$ and lasted approximately five weeks.

\setlength{\extrarowheight}{1.4pt}
\begin{table}
\centering
\scalebox{0.81}{
\begin{tabular}{l l l}
\hline  
 \textbf{Criteria} & \textbf{Question} & \textbf{Answer Options}  \\  \hline  
 Answer Confidence & \makecell[l]{How confident are you that your\\ provided annotations are correct?} & \makecell[l]{\textbf{1} (no confidence) \\ \textbf{5} (very confident that annotations are correct)}\\
 \hline
 Dataset Understanding & \makecell[l]{How well did you understand the\\ dataset (e.g. the task, domain, \\modality, etc.)?} & \makecell[l]{\textbf{1} (I don't understand the dataset at all) - \\\textbf{5} (the dataset incl. its purpose, creation, etc. \\is very clear and understandable for me)}\\
 \hline
 Completeness & \makecell[l]{Is there any (in your opinion \\important) information about the\\ dataset which you can’t define \\using Croissant?} & \makecell[l]{\textbf{1} (yes, there is lots of critical information about\\ the dataset that Croissant does not capture) - \\\textbf{5} (no, every important information about this \\dataset, which might be useful for ML users, \\is capture in Croissant attributes)}\\
 \hline
 Conciseness & \makecell[l]{Did you find any attributes redundant \\and not definable for this dataset?} & \makecell[l]{\textbf{1} (yes, there are lots of redundant attributes) - \\\textbf{5} (no, none of the attributes is redundant)}\\
 \hline
 Readability & \makecell[l]{How intuitive are the attributes names\\ for you? A name is not intuitive if\\ you need to check the specification\\ to understand the attribute's name?} & \makecell[l]{\textbf{1} (not intuitive at all, for each single attribute \\I checked the specification to understand it) - \\\textbf{5} (very intuitive, based on the name I could \\understand the attribute very well)}\\
 \hline
 Understandability & \makecell[l]{Rate the ease of understanding \\the Croissant specification.} & \makecell[l]{\textbf{1} (Understanding the spec. was very hard) - \\\textbf{5} (the spec. is very easy to understand)}\\
\hline
\end{tabular}}
\vspace{2mm}
\caption{\label{tab:evaluation_questions} Post-annotation assessment: Criteria, corresponding questions, and answer scales.}
\vspace{-3mm}
\end{table}

\paragraph{Selection of Croissant Attributes.}

We selected ten attributes from Croissant's Dataset Layer (see~\Cref{ssec:dataset_layer}) and six Croissant-RAI attributes (\Cref{ssec:croissant_rai}). 
We selected attributes that $(i)$ require manual specification, 
$(ii)$ can be defined by dataset users using the following resources: the dataset itself, a publication describing the dataset, and the Hugging Face dataset card if available, and 
$(iii)$ support the discoverability and reproducibility of datasets, along the lines of previous literature on improving dataset usability via documentation. For example, missing or limited descriptions of datasets reduce their discoverability and hinder practitioners from using the dataset as intended~\citep{YangLZ24}. Moreover, lack of information on data reproducibility, e.g., about the data collection and curation process, also impacts the dataset's adoption in the ML community~\citep{YangLZ24}. \Cref{tab:croissant_attributes} and \Cref{tab:croissant_rai_attributes} list the attributes selected for this study.

\paragraph{ML Datasets.}

We selected commonly used ML datasets from the language, vision, and audio modalities, based on their popularity on the Hugging Face (HF) Datasets repository. We further filtered datasets to require $(a)$ a pre-existing Croissant description, $(b)$ a dataset card in HF Datasets, and $(c)$ a publication that describes the dataset creation process. \Cref{tab:annotation_datasets} lists all datasets.

\paragraph{Evaluation.}
To evaluate the collected attribute annotations, we studied the provided answers and assessed the agreement among annotators. To measure agreement for textual attributes (e.g., \texttt{sc:description}), we calculated BLEU scores between attribute annotations, which were in textual form and did not allow for inter-annotator agreement scores commonly used for measuring agreement based on categorical data.
The BLEU metrics~\citep{papineni-etal-2002-bleu} measure text similarity based on overlapping $4$-grams in text pairs. The score can be between $[0, 1]$ with $1$ indicating perfect match between both compared texts.
Hence, a score closer to one indicates higher agreement among all three annotations available for the respective attribute and dataset.

\begin{table}
\centering
\begin{minipage}[t]{.20\linewidth}
\scalebox{0.80}{
\begin{tabular}[t]{l}
\hline  
 \textbf{Property}\\ \hline
 sc:description \\
sc:license \\
sc:name \\
sc:url \\
sc:creator \\
sc:publisher \\
sc:datePublished \\
sc:inLanguage \\
cr:citeAs \\
cr:isLiveDataset \\
\hline
\end{tabular}}
\vspace{2mm}
\caption{\label{tab:croissant_attributes} Annotated Croissant attributes.}
\end{minipage}%
\hspace{0.01\linewidth}
\begin{minipage}[t]{.45\linewidth}
\centering
\scalebox{0.80}{
\begin{tabular}[t]{l c}
\hline  
 \textbf{Property} & \textbf{RAI Use Case} \\ \hline
 rai:dataCollection & Data life cycle \\
 rai:dataCollectionTimeframe & Data life cycle\\
 rai:dataAnnotationPlatform & Data labelling\\
 rai:annotatorDemographics & Data labelling\\
 rai:dataUseCases & \makecell{AI safety and \\ fairness evaluation}\\
 rai:personalSensitiveInformation & Compliance\\ 
\hline
\end{tabular}}
\vspace{13mm}
\caption{\label{tab:croissant_rai_attributes} Annotated Croissant-RAI attributes.}
 \end{minipage} 
 \hspace{0.01\linewidth}
\begin{minipage}[t]{.30\linewidth}
\centering
\scalebox{0.80}{
\begin{tabular}[t]{l c}
\hline  
 \textbf{Dataset} & \textbf{Modality} \\ \hline
MMLU~\citep{HendrycksBBZMSS21}  & Language \\
Dolly-15k~\citep{DatabricksBlog2023DollyV2} & Language \\
FLORES~\citep{guzman-etal-2019-flores}  & Language \\
CIFAR10~\citep{Krizhevsky09learningmultiple} & Vision \\
MSCOCO~\citep{lin2014microsoft} & Vision \\
Visual Genome~\citep{Krishna2016VisualGC} & Vision \\
MMMU~\citep{yue2023mmmu} & VL \\
MathVista~\citep{lu2024mathvista} & VL \\
MLS\_Eng~\citep{Pratap2020MLSAL} & Audio \\
librispeech\_asr~\citep{panayotov2015librispeech} & Audio \\
\hline
\end{tabular}}
\vspace{2mm}
\caption{\label{tab:annotation_datasets} Annotated datasets.}
\end{minipage}
\vspace{-7mm}
\end{table}

\subsection{Mapping Evaluation Criteria to Croissant}

Previous literature proposes different criteria for vocabulary evaluation~\citep{Gomez-Perez01, HartmannBPG06}. Following prior work \citep{HartmannBPG06}, we evaluate Croissant on the five criteria we outline below. 
We further discuss how the criteria translate to Croissant and specify questions to evaluate each criterion in the context of our user study. 

\textbf{$(1)$ Consistency.} The criterion evaluates if a vocabulary is consistent and free of contradictions in its attribute definitions~\citep{Gomez-Perez01}. To measure Croissant's consistency, we studied how well annotations by different annotators for the same attribute and dataset aligned, i.e. based on the agreement among annotators.

\textbf{$(2)$ Completeness.} A vocabulary is complete if it covers the specified intent. 
While Croissant is an ongoing effort and not fully complete, we evaluated during the user study
if Croissant currently misses any attributes necessary to capture important information about commonly used ML datasets. We asked annotators to flag any important information about the datasets they annotated that could not be defined using Croissant. 

\textbf{$(3)$ Conciseness.} The conciseness criterion assesses whether a vocabulary avoids useless definitions and is free of redundancies. We measured this by asking annotators if they found any Croissant attributes redundant or not definable for the studied ML datasets. 

\textbf{$(4)$ Readability.}
The readability criteria assess how intuitive the attribute names are. After completing the annotations, we asked annotators to indicate on a Likert scale of $1$ to $5$ how intuitive they found Croissant attribute names to be.

\textbf{$(5)$ Understandability.}
The understandability criteria evaluates how easily user can understand Croissant attributes from the provided documentation. During our user study, we instructed annotators to use the Croissant specifications~\citep{croissant_spec, croissant_rai} and prompted them afterwards with questions.

\subsection{Results and Discussion}
\label{ssec:userstudy_results}

This section analyses data collected during the user study. First, we evaluate the answers to the questions listed in \Cref{tab:evaluation_questions}. Second, we study the annotation of Croissant and Croissant-RAI attributes.

\begin{figure}[h!] 
\begin{minipage}{.49\linewidth}
    \centering
    \includegraphics[width=1\columnwidth]{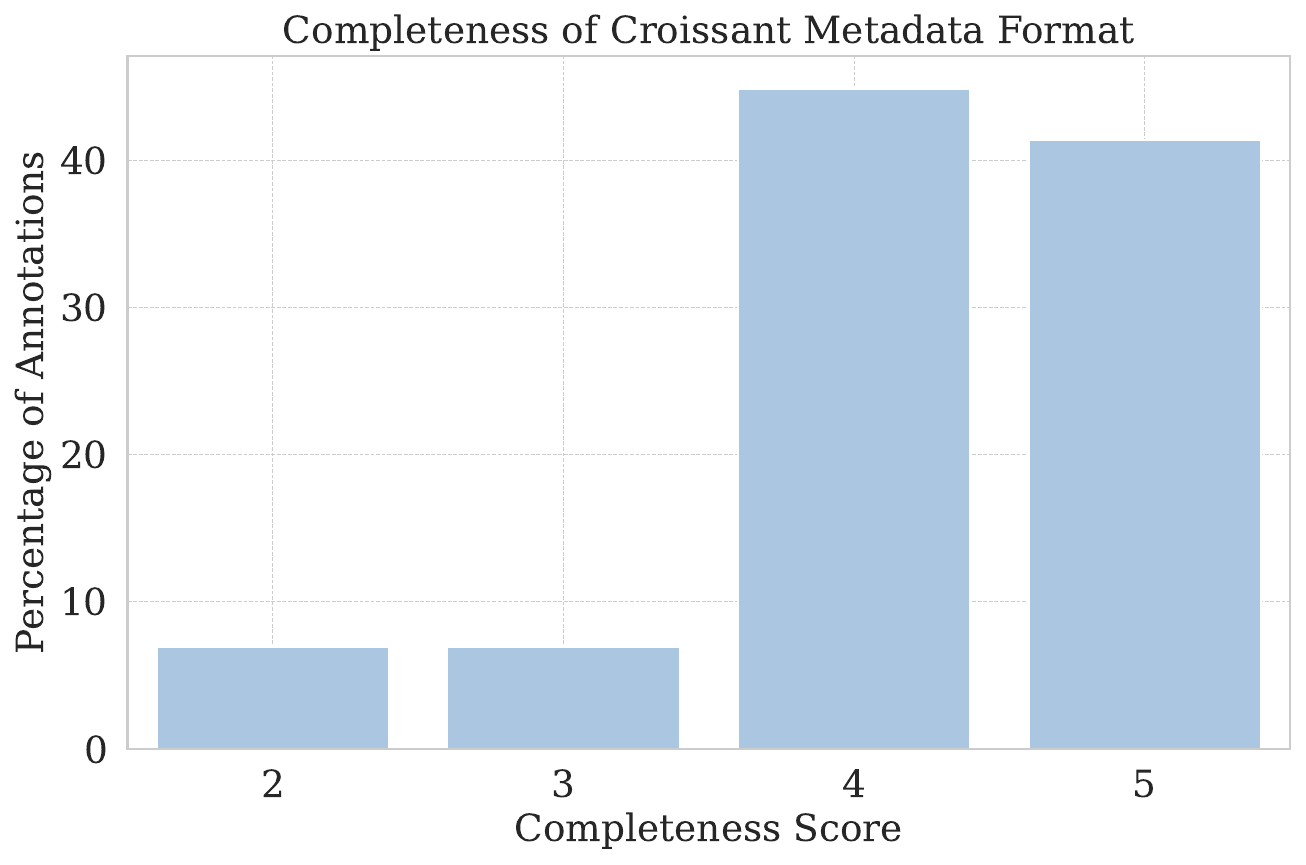}
    \caption{Answers to the \textit{completeness} question.}
    \label{fig:completeness}
\end{minipage}%
\hspace{0.01\linewidth}
\begin{minipage}{.49\linewidth}
    \centering
    \includegraphics[width=1\columnwidth]{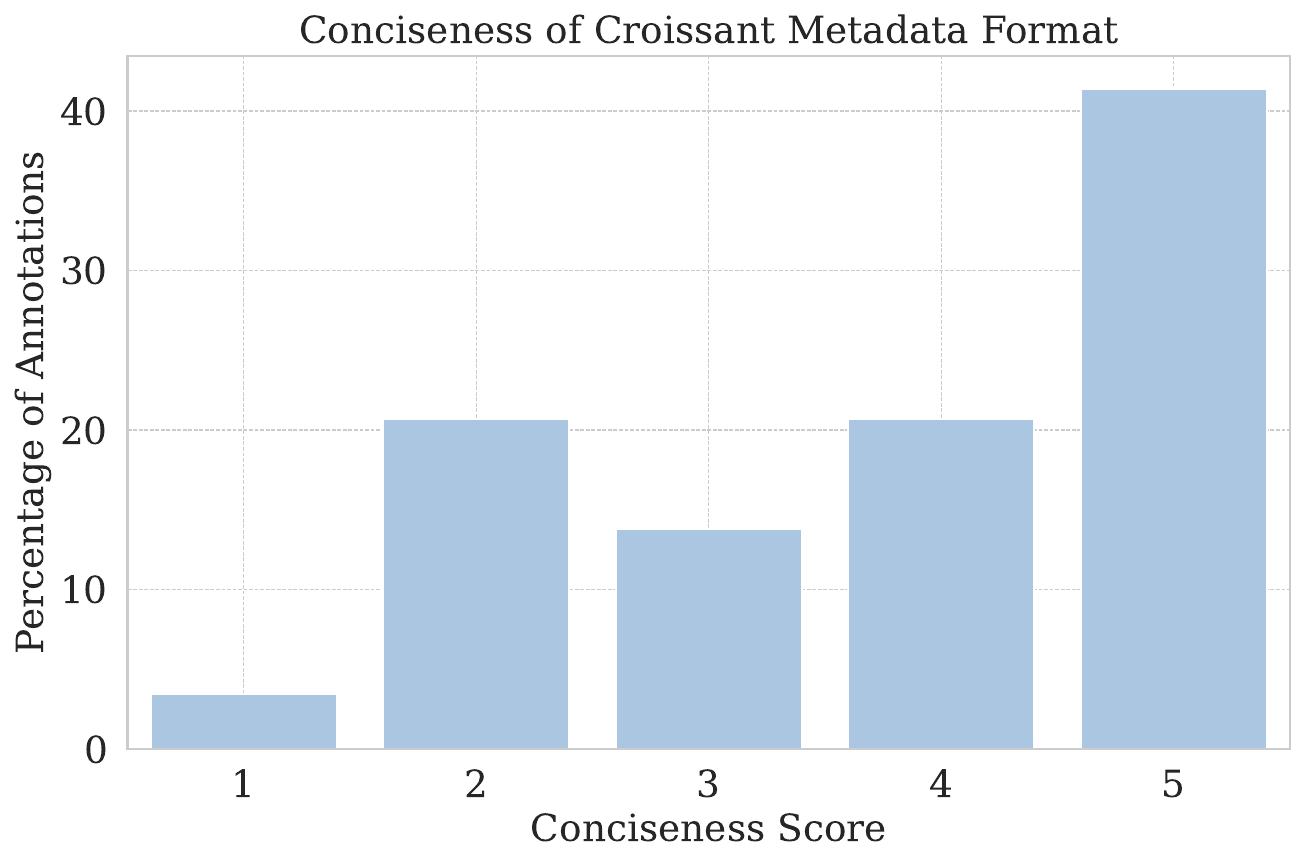}
    \caption{Answers to the \textit{conciseness} question.}
    \label{fig:conciseness}
\end{minipage}%
\vspace{-1mm}
\end{figure}

\begin{figure}[h!] 
\begin{minipage}{.49\linewidth}
    \centering
    \includegraphics[width=1\columnwidth]{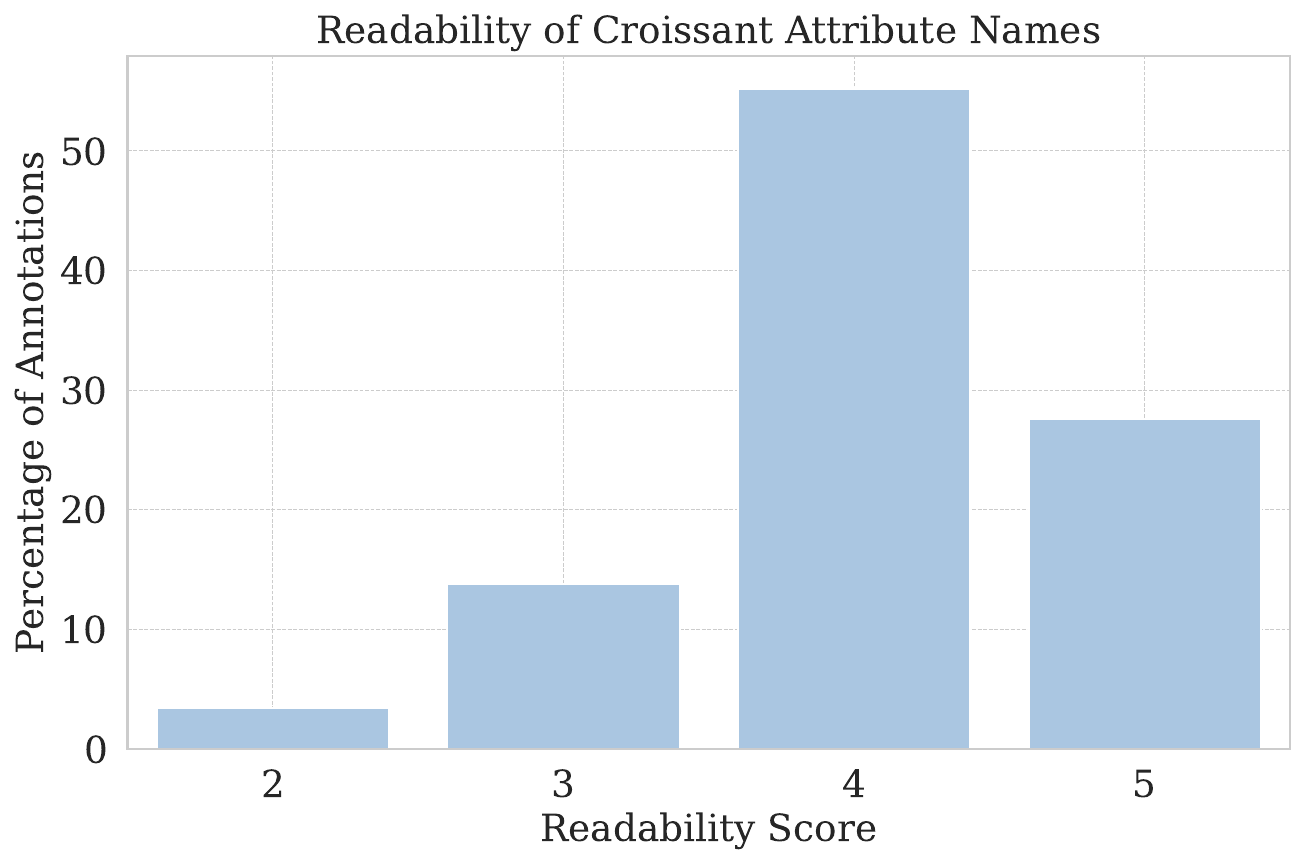}
    \caption{Answers to the \textit{readability} question.}
    \label{fig:readability}
\end{minipage}%
\hspace{0.01\linewidth}
\begin{minipage}{.49\linewidth}
    \centering
    \includegraphics[width=1\columnwidth]{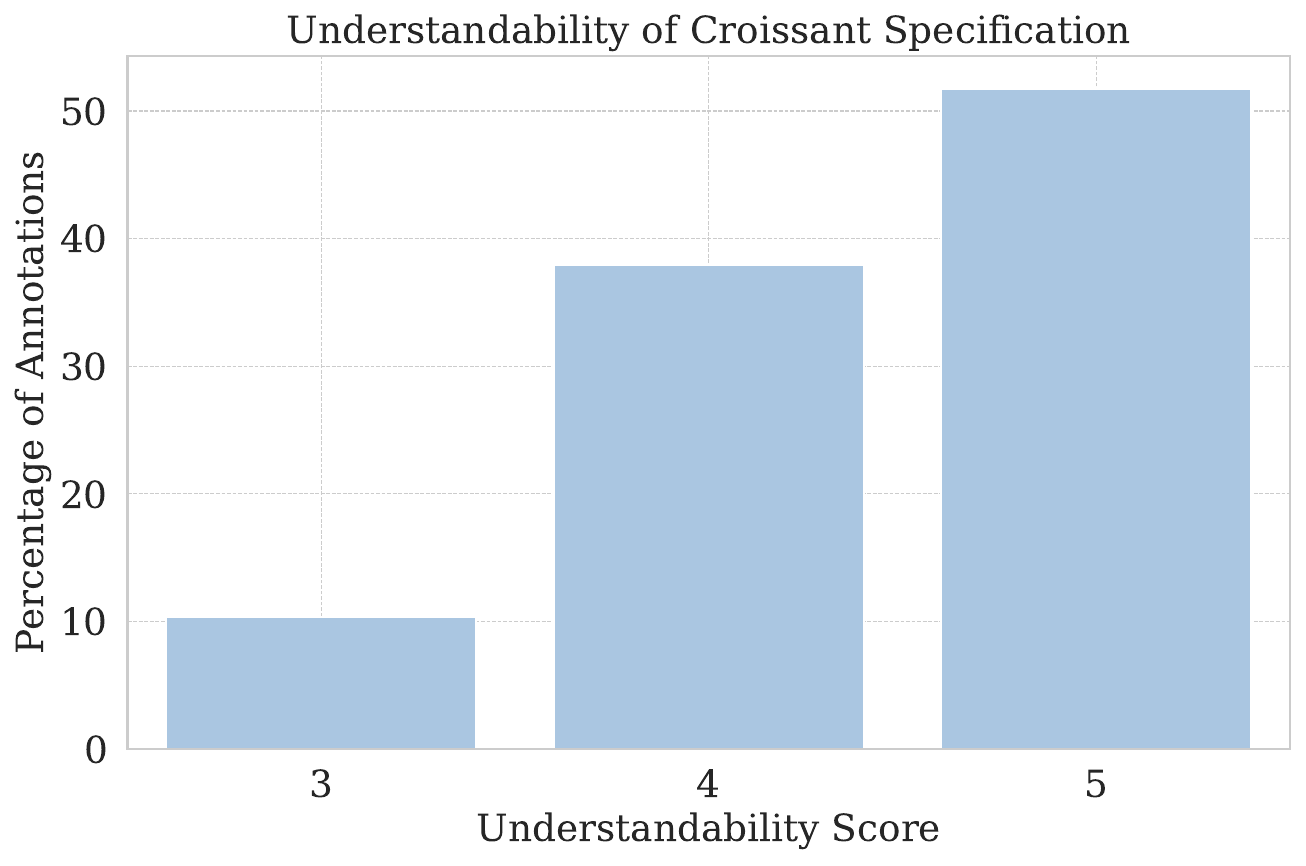}
    \caption{Answers to \textit{understandability} question.}
    \label{fig:understandability}
\end{minipage}%
\vspace{-3mm}
\end{figure}

\paragraph{Criteria Evaluation.}

Assessing annotators' ratings for the criteria in~\Cref{tab:evaluation_questions}, we find that for over $80\%$ of annotations ($25$ out of $30$ annotations), Croissant attributes capture important information about the datasets (see \Cref{fig:completeness}). 
For the conciseness criteria, we found a higher variance in ratings. While most annotations state that none or few requested attributes are redundant for the dataset (approx. $60\%$), seven annotations (around $23\%$) state that some attributes are redundant or not definable. This was due to Croissant-RAI attributes, e.g., \texttt{rai:annotatorDemographics} and \texttt{rai:personalSensitiveInformation}, which was either missing from the dataset's documentation or difficult to extract it.
The majority of annotations (around $83\%$) found Croissant and Croissant-RAI attribute names intuitive, resulting in high \textit{readability} scores (\Cref{fig:readability}). However, one attribute pair commonly confused annotators: \texttt{sc:creator} and \texttt{sc:publisher}. These attributes come from the schema.org vocabulary, which Croissant builds upon, and are already widely used to describe datasets on the Web.
On the bright side, around $90\%$ of annotations stated that the specifications were understandable(\Cref{fig:understandability}).

In addition to criteria-related questions, we asked annotators how confident they were regarding the correctness of their annotations and their understanding of the datasets. 
For the majority of annotations (more than $75\%$), their annotators selected that they were very confident. We found that the annotations with moderate confidence were indeed the ones with lower agreement on attribute values.
Moreover, for the majority of annotations, the annotators selected that they had a clear understanding of the datasets, with around $80\%$ selecting four or five on the five point scale (see \Cref{tab:evaluation_questions}), which gave us strong confidence in the data collected through this study. 

\paragraph{Attributes Evaluation.}
\Cref{tab:bleu_results} provides BLEU scores \cite{papineni-etal-2002-bleu} as a measure of agreement for annotated text attributes. Overall, the average BLEU scores for Croissant attributes ($0.55$) is higher than for Croissant-RAI attributes ($0.41$).
This difference can be attributed to several factors. First, multiple RAI attributes require a free-form text answer, which is more likely to differ across annotations than categorical or short-answer attributes such as \texttt{sc:name}, \texttt{sc:url}, or \texttt{sc:inLanguage}. Second, Croissant attributes are more easily extractable from the dataset's page on Hugging Face or from the introduction of the corresponding publication, while Croissant-RAI attributes often require detailed studying of the publication to find relevant RAI information, such as demographic information.

Attributes' average BLEU scores also diverges based on their expected values. Attributes with \texttt{Text} as the expected value have an average BLEU score of $0.4$ while \texttt{Date/Datetime} attributes have an average score of $0.47$.\footnote{See \url{https://schema.org/Text} and \url{https://schema.org/Date} for the exact definitions} Attributes with predefined values such as \texttt{Language} or \texttt{Url}, have an average score of $0.59$, indicating higher agreement.
For example, comparing annotations across attributes, we observe the highest BLEU scores for \texttt{sc:license}. This is largely attributable to the fact that, while being free-form text, there is less variety in the attribute's annotations and therefore more matching $4$-grams. The low BLEU score for the MathVista dataset is due to one annotator providing the text of the license instead of its name, as instructed in the specification~\citep{croissant_spec}.\footnote{For future Croissant versions, we plan to formalize some free text attributes.}

\begin{table}[t!]
\centering
\scalebox{0.68}{
\begin{tabular}{l c c c c c c c c c c c c c c}
\hline
 \textbf{Dataset} & \textbf{desc} & \textbf{lic} & \textbf{url} & \textbf{creator} & \textbf{publ} & \textbf{datePub} & \textbf{lang} & \textbf{citeAs} & \textbf{dataCol} & \textbf{time} & \textbf{plat} & \textbf{demogr} & \textbf{useCases} & \textbf{persInfo} \\ \hline
flores & 0.03 & 0.6 & 0.45 & 0.88 & 0.54 & 0.12 & 0.84 & 0.31 & 0.4 & 0.08 & 0.34 & 0.0 & 0.42 & 0.0 \\ 
cifar-10 & 0.39 & 1.0 & 0.31 & 0.17 & 0.16 & 0.14 & 1.0 & 0.26 & 0.35 & 0.0 & 1.0 & 0.0 & 0.29 & 1.0 \\ 
dolly-15k & 0.56 & 1.0 & 1.0 & 0.82 & 0.5 & 0.28 & 0.34 & 0.75 & 0.57 & 0.0 & 1.0 & 0.0 & 0.39 & 0.01 \\ 
mscoco & 0.7 & 1.0 & 0.65 & 0.26 & 0.0 & 0.24 & 1.0 & 0.0 & 0.32 & 1.0 & 0.78 & 0.0 & 0.88 & 0.0 \\ 
visual gen & 0.41 & 1.0 & 0.18 & 0.49 & 0.0 & 0.51 & 1.0 & 1.0 & 0.29 & 0.19 & 0.0 & 0.84 & 0.27 & 1.0 \\ 
mmmu & 0.89 & 0.49 & 1.0 & 0.76 & 0.33 & 0.21 & 1.0 & 1.0 & 0.77 & 1.0 & 0.0 & 0.05 & 0.48 & 0.62 \\ 
mmlu & 0.13 & 0.0 & 0.56 & 0.97 & 0.37 & 0.32 & 1.0 & 0.79 & 0.6 & 1.0 & 0.07 & 0.65 & 0.45 & 0.0 \\ 
mathvista & 1.0 & 0.34 & 0.57 & 0.53 & 0.07 & 0.26 & 0.13 & 1.0 & 0.16 & 1.0 & 0.0 & 0.05 & 0.22 & 1.0 \\ 
mls\_eng & 0.35 & 1.0 & 1.0 & 0.64 & 0.35 & 0.56 & 0.03 & 1.0 & 0.3 & 1.0 & 0.0 & 1.0 & 0.36 & 0.0 \\ 
librispeech & 0.73 & 1.0 & 0.17 & 0.82 & 0.33 & 0.44 & 1.0 & 0.04 & 0.34 & 1.0 & 0.0 & 0.29 & 0.25 & 0.21 \\
\hline
Average & 0.52 & 0.74 & 0.59 & 0.63 & 0.26 & 0.31 & 0.73 & 0.62 & 0.41 & 0.63 & 0.32 & 0.29 & 0.4 & 0.38 \\ 
Median & 0.52 & 1.0 & 0.57 & 0.64 & 0.33 & 0.28 & 1.0 & 0.75 & 0.35 & 1.0 & 0.07 & 0.05 & 0.39 & 0.21 \\ 
\hline
\end{tabular}}
\vspace{2mm}
\caption{\label{tab:bleu_results} BLEU scores for annotated datasets and attributes (i.e. \textbf{desc}ription, \textbf{lic}ense, url, creator, \textbf{publ}isher, \textbf{datePub}lished, in\textbf{Lang}uage, citeAs, \textbf{dataColl}ection, dataCollection\textbf{Time}frame, dataAnnotation\textbf{Plat}form, annotator\textbf{Demogr}aphics, data\textbf{UseCases}, \textbf{pers}onalSensitive\textbf{Info}rmation)}
\vspace{-7mm}
\end{table}

\section{Limitations and Future Work}%

\paragraph{Croissant Format.}

While the Croissant metadata format provides a shared representation across various ML tools, platforms, and frameworks, certain challenges remain that should be addressed in future work. First, its structure may pose difficulties for users unfamiliar with the format, potentially hindering broader adoption. In the future, we plan to extend Croissant tools (e.g., the Croissant editor) and provide comprehensive documentation, as these are the primary means of making Croissant datasets easier for users to utilize. This includes adding additional annotated dataset examples in the Croissant repository and developing community guidelines that account for domain-specific needs.
Second, the Croissant editor, as an interface for users to create Croissant metadata, is still in its early stages and will be further developed and enhanced in future work. For example, it will support additional functionalities such as archiving files, nested fields, and more.
Finally, as part of an ongoing effort to enhance and demonstrate Croissant's handling of complex data structures, a GeoSpatial extension for Croissant (named Geo-Croissant) will be released in the near future, accompanied by examples that demonstrate the handling of file formats such as HDF5 and Zarr.

\paragraph{User Study.}

While the user study presented here provides some initial insights on the usability of the Croissant vocabulary, it has a number of limitations that warrant follow-on work: $(i)$ Increase the number of participants and annotated datasets, either by recruiting participants with the right combination of skills in dataset documentation and machine learning, or by encouraging the authors of datasets to create annotations directly, as they are most knowledgeable. 
Moreover, as the annotators were drawn from the Croissant community, this may have introduced bias as they were are familiar with the Croissant framework to some extent. Hence, a required future direction is exploring Croissant usage by non-expert users.
$(ii)$ Extend the evaluation results, as participants only focused on a subset of the attributes in the Croissant and Croissant RAI vocabularies. $(iii)$ Finally, BLEU scores are a noisy metric for the quality of attribute annotations. However, due to the nature of our collected data being textual rather than categorical, standard agreement metrics such as Fleiss' Kappa~\cite{fleiss1971measuring} were not applicable.
A possible future direction is to compare annotations with golden data from existing Croissant descriptions for the datasets. 
\section{Conclusion}%

This paper introduces Croissant, a metadata format for ML datasets. Croissant improves the discoverability, portability, and interoperability of ML datasets across data repositories, ML tools, frameworks, and platforms. The Croissant format addresses key challenges in data management by providing a standardized data representation, making datasets more discoverable, portable, and interoperable. 

Croissant has already seen rapid adoption by popular ML dataset repositories and frameworks, and is recommended as a data artifact in the NeurIPS Datasets and Benchmarks Track. Moreover, Croissant metadata is deemed readable, understandable, complete, and concise by human raters. Still, Croissant's success will ultimately depend on further adoption in ML research and industry, the widespread availability of Croissant datasets, and support from ML tools and frameworks, thus we warmly invite further ML platform and tool developers to join the Croissant community.

Finally, Croissant's extendable nature and the broad range of datasets it can represent enables
other communities to extend Croissant for their specific needs, similar to the Croissant-RAI extension developed for Responsible AI, and streamline collaborations between ML and other fields.

\newpage%
\section*{Acknowledgments}

This work was partly funded by the HE project MuseIT, which has been co-founded by the European Union under the Grant Agreement No 101061441. MuseIT has supported the work of Nitisha Jain. Views and opinions expressed are, however, those of the authors and do not necessarily reflect those of the European Union or European Research Executive Agency. Joan Giner-Miguelez is supported by the AIDOaRt project, which is funded by the ECSEL Joint Undertaking (JU) under grant agreement No 101007350. The JU receives support from the European Union’s Horizon 2020 research and innovation programme and Sweden, Austria, Czech Republic, Finland, France, Italy, and Spain. Pieter
Gijsbers, Joaquin Vanschoren, and Jos van der Velde would like to acknowledge funding by EU’s Horizon Europe research and innovation program under grant agreement No. 952215 (TAILOR) and No. 101070000 (AI4EUROPE).
\bibliographystyle{unsrtnat}
\bibliography{references}%

\begin{thebibliography}{49}
\providecommand{\natexlab}[1]{#1}
\providecommand{\url}[1]{\texttt{#1}}
\expandafter\ifx\csname urlstyle\endcsname\relax
  \providecommand{\doi}[1]{doi: #1}\else
  \providecommand{\doi}{doi: \begingroup \urlstyle{rm}\Url}\fi

\bibitem[Kuchnik et~al.(2022)Kuchnik, Klimovic, Simsa, Smith, and Amvrosiadis]{kuchnik2022plumber}
Michael Kuchnik, Ana Klimovic, Jiri Simsa, Virginia Smith, and George Amvrosiadis.
\newblock Plumber: Diagnosing and removing performance bottlenecks in machine learning data pipelines.
\newblock \emph{Proceedings of Machine Learning and Systems}, 4:\penalty0 33--51, 2022.

\bibitem[Oala et~al.(2024)Oala, Maskey, Bat-Leah, Parrish, G{\"u}rel, Kuo, Liu, Dror, Brajovic, Yao, Bartolo, Rojas, Hileman, Aliment, Mahoney, Risdal, Lease, Samek, Dutta, Northcutt, Coleman, Hancock, Koch, Tadesse, Karla{\v{s}}, Alaa, Dieng, Noy, Reddi, Zou, Paritosh, van~der Schaar, Bollacker, Aroyo, Zhang, Vanschoren, Guyon, and Mattson]{2024dmlr}
Luis Oala, Manil Maskey, Lilith Bat-Leah, Alicia Parrish, Nezihe~Merve G{\"u}rel, Tzu-Sheng Kuo, Yang Liu, Rotem Dror, Danilo Brajovic, Xiaozhe Yao, Max Bartolo, William A~Gaviria Rojas, Ryan Hileman, Rainier Aliment, Michael~W. Mahoney, Meg Risdal, Matthew Lease, Wojciech Samek, Debojyoti Dutta, Curtis~G Northcutt, Cody Coleman, Braden Hancock, Bernard Koch, Girmaw~Abebe Tadesse, Bojan Karla{\v{s}}, Ahmed Alaa, Adji~Bousso Dieng, Natasha Noy, Vijay~Janapa Reddi, James Zou, Praveen Paritosh, Mihaela van~der Schaar, Kurt Bollacker, Lora Aroyo, Ce~Zhang, Joaquin Vanschoren, Isabelle Guyon, and Peter Mattson.
\newblock {DMLR}: Data-centric machine learning research - past, present and future.
\newblock \emph{Journal of Data-centric Machine Learning Research}, 2024.
\newblock URL \url{https://openreview.net/forum?id=2kpu78QdeE}.
\newblock Featured Certification, Survey Certification.

\bibitem[Sambasivan et~al.(2021)Sambasivan, Kapania, Highfill, Akrong, Paritosh, and Aroyo]{sambasivan2021everyone}
Nithya Sambasivan, Shivani Kapania, Hannah Highfill, Diana Akrong, Praveen Paritosh, and Lora~M Aroyo.
\newblock “{Everyone} wants to do the model work, not the data work”: {Data Cascades in High-Stakes AI}.
\newblock In \emph{proceedings of the 2021 CHI Conference on Human Factors in Computing Systems}, pages 1--15, 2021.

\bibitem[Akhtar et~al.(2024{\natexlab{a}})Akhtar, Benjelloun, Conforti, Gijsbers, Giner-Miguelez, Jain, Kuchnik, Lhoest, Marcenac, Maskey, Mattson, Oala, Ruyssen, Shinde, Simperl, Thomas, Tykhonov, Vanschoren, van~der Velde, Vogler, and Wu]{10.1145/3650203.3663326}
Mubashara Akhtar, Omar Benjelloun, Costanza Conforti, Pieter Gijsbers, Joan Giner-Miguelez, Nitisha Jain, Michael Kuchnik, Quentin Lhoest, Pierre Marcenac, Manil Maskey, Peter Mattson, Luis Oala, Pierre Ruyssen, Rajat Shinde, Elena Simperl, Goeffry Thomas, Slava Tykhonov, Joaquin Vanschoren, Jos van~der Velde, Steffen Vogler, and Carole-Jean Wu.
\newblock Croissant: A metadata format for ml-ready datasets.
\newblock In \emph{Proceedings of the Eighth Workshop on Data Management for End-to-End Machine Learning}, DEEM '24, page 1–6, New York, NY, USA, 2024{\natexlab{a}}. Association for Computing Machinery.
\newblock ISBN 9798400706110.
\newblock \doi{10.1145/3650203.3663326}.
\newblock URL \url{https://doi.org/10.1145/3650203.3663326}.

\bibitem[Benjelloun et~al.(2024)Benjelloun, Simperl, Marcenac, Ruyssen, Conforti, Kuchnik, van~der Velde, Oala, Vogler, Akhtar, Jain, and Tykhonov]{croissant_spec}
Omar Benjelloun, Elena Simperl, Pierre Marcenac, Pierre Ruyssen, Costanza Conforti, Michael Kuchnik, Jos van~der Velde, Luis Oala, Steffen Vogler, Mubashara Akhtar, Nitisha Jain, and Slava Tykhonov.
\newblock Croissant format specification.
\newblock Technical report, 2024.
\newblock URL \url{https://mlcommons.org/croissant/1.0}.

\bibitem[Gebru et~al.(2021)Gebru, Morgenstern, Vecchione, Vaughan, Wallach, III, and Crawford]{gebru2021datasheets}
Timnit Gebru, Jamie Morgenstern, Briana Vecchione, Jennifer~Wortman Vaughan, Hanna Wallach, Hal~Daumé III, and Kate Crawford.
\newblock Datasheets for datasets, 2021.

\bibitem[Wilson et~al.(2022)Wilson, Goonetillake, Indika, and Ginige]{wilson2022conceptual}
RSI Wilson, JS~Goonetillake, WA~Indika, and Athula Ginige.
\newblock A conceptual model for ontology quality assessment.
\newblock \emph{Semantic Web}, \penalty0 (Preprint):\penalty0 1--47, 2022.

\bibitem[Albertoni et~al.(2024)Albertoni, Browning, Cox, Beltran, Perego, and Winstanley]{DCAT}
Riccardo Albertoni, David Browning, Simon J~D Cox, Alejandra~Gonzalez Beltran, Andrea Perego, and Peter Winstanley.
\newblock Data catalog vocabulary ({DCAT}) - version 3.
\newblock \url{https://www.w3.org/TR/vocab-dcat-3/}, 01 2024.
\newblock (Accessed on 03/18/2024).

\bibitem[schema.org(2024)]{schema_dataset}
schema.org.
\newblock Schema.org v26.0.
\newblock \url{https://github.com/schemaorg/schemaorg/tree/main/data/releases/26.0/}, 02 2024.
\newblock (Accessed on 03/18/2024).

\bibitem[Guha et~al.(2016)Guha, Brickley, and Macbeth]{guha2016schema}
Ramanathan~V Guha, Dan Brickley, and Steve Macbeth.
\newblock Schema. org: evolution of structured data on the web.
\newblock \emph{Communications of the ACM}, 59\penalty0 (2):\penalty0 44--51, 2016.

\bibitem[Group(2024{\natexlab{a}})]{data-packages}
Frictionless~Working Group.
\newblock Data packages.
\newblock \url{https://specs.frictionlessdata.io/}, 2024{\natexlab{a}}.
\newblock (Accessed on 03/21/2024).

\bibitem[Group(2016)]{csvw}
W3C~Working Group.
\newblock {CSV} on the web: A primer.
\newblock \url{https://www.w3.org/TR/tabular-data-primer/}, 2016.
\newblock (Accessed on 03/21/2024).

\bibitem[Lawson et~al.(2021)Lawson, Cabili, Kerry, Boughtwood, Thorogood, Alper, Bowers, Boyles, Brookes, Brush, et~al.]{lawson2021data}
Jonathan Lawson, Moran~N Cabili, Giselle Kerry, Tiffany Boughtwood, Adrian Thorogood, Pinar Alper, Sarion~R Bowers, Rebecca~R Boyles, Anthony~J Brookes, Matthew Brush, et~al.
\newblock The data use ontology to streamline responsible access to human biomedical datasets.
\newblock \emph{Cell Genomics}, 1\penalty0 (2), 2021.

\bibitem[Wilkinson et~al.(2016)Wilkinson, Dumontier, Aalbersberg, Appleton, Axton, Baak, Blomberg, Boiten, da~Silva~Santos, Bourne, et~al.]{wilkinson2016fair}
Mark~D Wilkinson, Michel Dumontier, IJsbrand~Jan Aalbersberg, Gabrielle Appleton, Myles Axton, Arie Baak, Niklas Blomberg, Jan-Willem Boiten, Luiz~Bonino da~Silva~Santos, Philip~E Bourne, et~al.
\newblock The fair guiding principles for scientific data management and stewardship.
\newblock \emph{Scientific data}, 3\penalty0 (1):\penalty0 1--9, 2016.

\bibitem[Jarrahi et~al.(2023)Jarrahi, Memariani, and Guha]{data-centric}
Mohammad~Hossein Jarrahi, Ali Memariani, and Shion Guha.
\newblock The principles of data-centric ai.
\newblock \emph{Commun. ACM}, 66\penalty0 (8):\penalty0 84–92, jul 2023.
\newblock ISSN 0001-0782.
\newblock \doi{10.1145/3571724}.
\newblock URL \url{https://doi.org/10.1145/3571724}.

\bibitem[Smuha(2019)]{Smuha2019}
Nathalie~A. Smuha.
\newblock {The EU Approach to Ethics Guidelines for Trustworthy Artificial Intelligence}.
\newblock \emph{Computer Law Review International}, 20\penalty0 (4):\penalty0 97--106, 2019.
\newblock \doi{doi:10.9785/cri-2019-200402}.
\newblock URL \url{https://doi.org/10.9785/cri-2019-200402}.

\bibitem[Bender and Friedman(2018)]{bender-friedman-2018-data}
Emily~M. Bender and Batya Friedman.
\newblock Data statements for natural language processing: Toward mitigating system bias and enabling better science.
\newblock \emph{Transactions of the Association for Computational Linguistics}, 6:\penalty0 587--604, 2018.
\newblock \doi{10.1162/tacl_a_00041}.
\newblock URL \url{https://aclanthology.org/Q18-1041}.

\bibitem[Pushkarna et~al.(2022)Pushkarna, Zaldivar, and Kjartansson]{pushkarna2022data}
Mahima Pushkarna, Andrew Zaldivar, and Oddur Kjartansson.
\newblock Data cards: Purposeful and transparent dataset documentation for responsible ai, 2022.

\bibitem[Holland et~al.(2018)Holland, Hosny, Newman, Joseph, and Chmielinski]{holland2018dataset}
Sarah Holland, Ahmed Hosny, Sarah Newman, Joshua Joseph, and Kasia Chmielinski.
\newblock The dataset nutrition label: A framework to drive higher data quality standards, 2018.

\bibitem[Kaggle(2024)]{kaggle_datasets}
Kaggle.
\newblock Kaggle datasets: A platform for data science competitions and collaborative work, 2024.
\newblock URL \url{https://www.kaggle.com/datasets}.

\bibitem[Vanschoren et~al.(2013)Vanschoren, van Rijn, Bischl, and Torgo]{OpenML2013}
Joaquin Vanschoren, Jan~N. van Rijn, Bernd Bischl, and Luis Torgo.
\newblock {OpenML}: networked science in machine learning.
\newblock \emph{SIGKDD Explorations}, 15\penalty0 (2):\penalty0 49--60, 2013.
\newblock \doi{10.1145/2641190.2641198}.
\newblock URL \url{http://doi.acm.org/10.1145/2641190.264119}.

\bibitem[HuggingFace(2024{\natexlab{a}})]{huggingface_datasets}
HuggingFace.
\newblock {Hugging Face Datasets}: A community-driven hub for ready-to-use datasets, 2024{\natexlab{a}}.
\newblock URL \url{https://huggingface.co/datasets}.

\bibitem[HuggingFace(2024{\natexlab{b}})]{hf_datasets_cards}
HuggingFace.
\newblock {Hugging Face} dataset cards.
\newblock \url{https://huggingface.co/docs/hub/en/datasets-cards}, 2024{\natexlab{b}}.
\newblock (Accessed on 06/05/2024).

\bibitem[Xiao et~al.(2017)Xiao, Rasul, and Vollgraf]{fashion-mnist}
Han Xiao, Kashif Rasul, and Roland Vollgraf.
\newblock Fashion-mnist: a novel image dataset for benchmarking machine learning algorithms.
\newblock 2017.

\bibitem[Asano et~al.(2021)Asano, Rupprecht, Zisserman, and Vedaldi]{asano21pass}
Yuki~M. Asano, Christian Rupprecht, Andrew Zisserman, and Andrea Vedaldi.
\newblock Pass: An imagenet replacement for self-supervised pretraining without humans.
\newblock \emph{NeurIPS Track on Datasets and Benchmarks}, 2021.

\bibitem[Lin et~al.(2015)Lin, Maire, Belongie, Bourdev, Girshick, Hays, Perona, Ramanan, Zitnick, and Dollár]{coco}
Tsung-Yi Lin, Michael Maire, Serge Belongie, Lubomir Bourdev, Ross Girshick, James Hays, Pietro Perona, Deva Ramanan, C.~Lawrence Zitnick, and Piotr Dollár.
\newblock Microsoft coco: Common objects in context, 2015.

\bibitem[Akhtar et~al.(2024{\natexlab{b}})Akhtar, Jain, Giner-Miguelez, Benjelloun, Simperl, Aroyo, Shinde, Oala, and Kuchnik]{croissant_rai}
Mubashara Akhtar, Nitisha Jain, Joan Giner-Miguelez, Omar Benjelloun, Elena Simperl, Lora Aroyo, Rajat Shinde, Luis Oala, and Michael Kuchnik.
\newblock {Croissant RAI Specification}.
\newblock Technical report, 2024{\natexlab{b}}.
\newblock URL \url{https://mlcommons.org/croissant/RAI/1.0}.

\bibitem[Jain et~al.(2024)Jain, Akhtar, Giner-Miguelez, Shinde, Vanschoren, Vogler, Goswami, Rao, Santos, Oala, Karamousadakis, Maskey, Marcenac, Conforti, Kuchnik, Aroyo, Benjelloun, and Simperl]{Jain24}
Nitisha Jain, Mubashara Akhtar, Joan Giner-Miguelez, Rajat Shinde, Joaquin Vanschoren, Steffen Vogler, Sujata Goswami, Yuhan Rao, Tim Santos, Luis Oala, Michalis Karamousadakis, Manil Maskey, Pierre Marcenac, Costanza Conforti, Michael Kuchnik, Lora Aroyo, Omar Benjelloun, and Elena Simperl.
\newblock {A Standardized Machine-readable Dataset Documentation Format for Responsible AI}.
\newblock \emph{to appear in ArXiv}, abs/5643361, 2024.

\bibitem[Brickley et~al.(2019)Brickley, Burgess, and Noy]{dataset_search}
Dan Brickley, Matthew Burgess, and Natasha Noy.
\newblock Google dataset search: Building a search engine for datasets in an open web ecosystem.
\newblock In \emph{The world wide web conference}, pages 1365--1375, 2019.

\bibitem[TFDS(2024)]{TFDS}
TFDS.
\newblock {TensorFlow Datasets}, a collection of ready-to-use datasets.
\newblock \url{https://www.tensorflow.org/datasets}, 03 2024.

\bibitem[PyTorch(2024)]{pytorch_datapipes}
PyTorch.
\newblock {DataPipe Tutorial}.
\newblock \url{https://pytorch.org/data/beta/dp_tutorial.html}, 2024.
\newblock (Accessed on 10/28/2024).

\bibitem[G{\'{o}}mez{-}P{\'{e}}rez(2001)]{Gomez-Perez01}
Asunci{\'{o}}n G{\'{o}}mez{-}P{\'{e}}rez.
\newblock Evaluation of ontologies.
\newblock \emph{Int. J. Intell. Syst.}, 16\penalty0 (3):\penalty0 391--409, 2001.

\bibitem[Group(2024{\natexlab{b}})]{croissant_working_group_2024_13350974}
Croissant~Working Group.
\newblock Croissant - user research report, August 2024{\natexlab{b}}.
\newblock URL \url{https://doi.org/10.5281/zenodo.13350974}.

\bibitem[Likert(1932)]{likert1932technique}
Rensis Likert.
\newblock A technique for the measurement of attitudes.
\newblock \emph{Archives of Psychology}, 22\penalty0 (140):\penalty0 1--55, 1932.

\bibitem[Troiano et~al.(2021)Troiano, Pad{\'o}, and Klinger]{troiano-etal-2021-emotion}
Enrica Troiano, Sebastian Pad{\'o}, and Roman Klinger.
\newblock Emotion ratings: How intensity, annotation confidence and agreements are entangled.
\newblock In Orphee De~Clercq, Alexandra Balahur, Joao Sedoc, Valentin Barriere, Shabnam Tafreshi, Sven Buechel, and Veronique Hoste, editors, \emph{Proceedings of the Eleventh Workshop on Computational Approaches to Subjectivity, Sentiment and Social Media Analysis}, pages 40--49, Online, April 2021. Association for Computational Linguistics.
\newblock URL \url{https://aclanthology.org/2021.wassa-1.5}.

\bibitem[Yang et~al.(2024)Yang, Liang, and Zou]{YangLZ24}
Xinyu Yang, Weixin Liang, and James Zou.
\newblock Navigating dataset documentations in {AI:} {A} large-scale analysis of dataset cards on hugging face.
\newblock \emph{CoRR}, abs/2401.13822, 2024.
\newblock \doi{10.48550/ARXIV.2401.13822}.
\newblock URL \url{https://doi.org/10.48550/arXiv.2401.13822}.

\bibitem[Papineni et~al.(2002)Papineni, Roukos, Ward, and Zhu]{papineni-etal-2002-bleu}
Kishore Papineni, Salim Roukos, Todd Ward, and Wei-Jing Zhu.
\newblock {B}leu: a method for automatic evaluation of machine translation.
\newblock In Pierre Isabelle, Eugene Charniak, and Dekang Lin, editors, \emph{Proceedings of the 40th Annual Meeting of the Association for Computational Linguistics}, pages 311--318, Philadelphia, Pennsylvania, USA, July 2002. Association for Computational Linguistics.
\newblock \doi{10.3115/1073083.1073135}.
\newblock URL \url{https://aclanthology.org/P02-1040}.

\bibitem[Hendrycks et~al.(2021)Hendrycks, Burns, Basart, Zou, Mazeika, Song, and Steinhardt]{HendrycksBBZMSS21}
Dan Hendrycks, Collin Burns, Steven Basart, Andy Zou, Mantas Mazeika, Dawn Song, and Jacob Steinhardt.
\newblock Measuring massive multitask language understanding.
\newblock In \emph{9th International Conference on Learning Representations, {ICLR} 2021, Virtual Event, Austria, May 3-7, 2021}. OpenReview.net, 2021.
\newblock URL \url{https://openreview.net/forum?id=d7KBjmI3GmQ}.

\bibitem[Conover et~al.(2023)Conover, Hayes, Mathur, Xie, Wan, Shah, Ghodsi, Wendell, Zaharia, and Xin]{DatabricksBlog2023DollyV2}
Mike Conover, Matt Hayes, Ankit Mathur, Jianwei Xie, Jun Wan, Sam Shah, Ali Ghodsi, Patrick Wendell, Matei Zaharia, and Reynold Xin.
\newblock Free dolly: Introducing the world's first truly open instruction-tuned llm, 2023.
\newblock URL \url{https://www.databricks.com/blog/2023/04/12/dolly-first-open-commercially-viable-instruction-tuned-llm}.

\bibitem[Guzm{\'a}n et~al.(2019)Guzm{\'a}n, Chen, Ott, Pino, Lample, Koehn, Chaudhary, and Ranzato]{guzman-etal-2019-flores}
Francisco Guzm{\'a}n, Peng-Jen Chen, Myle Ott, Juan Pino, Guillaume Lample, Philipp Koehn, Vishrav Chaudhary, and Marc{'}Aurelio Ranzato.
\newblock The {FLORES} evaluation datasets for low-resource machine translation: {N}epali{--}{E}nglish and {S}inhala{--}{E}nglish.
\newblock In Kentaro Inui, Jing Jiang, Vincent Ng, and Xiaojun Wan, editors, \emph{Proceedings of the 2019 Conference on Empirical Methods in Natural Language Processing and the 9th International Joint Conference on Natural Language Processing (EMNLP-IJCNLP)}, pages 6098--6111, Hong Kong, China, November 2019. Association for Computational Linguistics.
\newblock \doi{10.18653/v1/D19-1632}.
\newblock URL \url{https://aclanthology.org/D19-1632}.

\bibitem[Krizhevsky(2009)]{Krizhevsky09learningmultiple}
Alex Krizhevsky.
\newblock Learning multiple layers of features from tiny images.
\newblock Technical report, 2009.

\bibitem[Lin et~al.(2014)Lin, Maire, Belongie, Hays, Perona, Ramanan, Doll{\'a}r, and Zitnick]{lin2014microsoft}
Tsung-Yi Lin, Michael Maire, Serge Belongie, James Hays, Pietro Perona, Deva Ramanan, Piotr Doll{\'a}r, and C~Lawrence Zitnick.
\newblock Microsoft coco: Common objects in context.
\newblock In \emph{Computer Vision--ECCV 2014: 13th European Conference, Zurich, Switzerland, September 6-12, 2014, Proceedings, Part V 13}, pages 740--755. Springer, 2014.

\bibitem[Krishna et~al.(2017)Krishna, Zhu, Groth, Johnson, Hata, Kravitz, Chen, Kalantidis, Li, Shamma, Bernstein, and Fei-Fei]{Krishna2016VisualGC}
Ranjay Krishna, Yuke Zhu, Oliver Groth, Justin Johnson, Kenji Hata, Joshua Kravitz, Stephanie Chen, Yannis Kalantidis, Li-Jia Li, David~A. Shamma, Michael~S. Bernstein, and Li~Fei-Fei.
\newblock Visual genome: Connecting language and vision using crowdsourced dense image annotations.
\newblock \emph{International Journal of Computer Vision}, 123:\penalty0 32--73, 2017.
\newblock \doi{10.1007/s11263-016-0981-7}.
\newblock URL \url{https://doi.org/10.1007/s11263-016-0981-7}.

\bibitem[Yue et~al.(2024)Yue, Ni, Zhang, Zheng, Liu, Zhang, Stevens, Jiang, Ren, Sun, Wei, Yu, Yuan, Sun, Yin, Zheng, Yang, Liu, Huang, Sun, Su, and Chen]{yue2023mmmu}
Xiang Yue, Yuansheng Ni, Kai Zhang, Tianyu Zheng, Ruoqi Liu, Ge~Zhang, Samuel Stevens, Dongfu Jiang, Weiming Ren, Yuxuan Sun, Cong Wei, Botao Yu, Ruibin Yuan, Renliang Sun, Ming Yin, Boyuan Zheng, Zhenzhu Yang, Yibo Liu, Wenhao Huang, Huan Sun, Yu~Su, and Wenhu Chen.
\newblock {MMMU: A Massive Multi-discipline Multimodal Understanding and Reasoning Benchmark for Expert AGI}.
\newblock In \emph{Proceedings of CVPR}, 2024.

\bibitem[Lu et~al.(2024)Lu, Bansal, Xia, Liu, Li, Hajishirzi, Cheng, Chang, Galley, and Gao]{lu2024mathvista}
Pan Lu, Hritik Bansal, Tony Xia, Jiacheng Liu, Chunyuan Li, Hannaneh Hajishirzi, Hao Cheng, Kai-Wei Chang, Michel Galley, and Jianfeng Gao.
\newblock Mathvista: Evaluating mathematical reasoning of foundation models in visual contexts.
\newblock In \emph{International Conference on Learning Representations (ICLR)}, 2024.

\bibitem[Pratap et~al.(2020)Pratap, Xu, Sriram, Synnaeve, and Collobert]{Pratap2020MLSAL}
Vineel Pratap, Qiantong Xu, Anuroop Sriram, Gabriel Synnaeve, and Ronan Collobert.
\newblock Mls: A large-scale multilingual dataset for speech research.
\newblock \emph{ArXiv}, abs/2012.03411, 2020.

\bibitem[Panayotov et~al.(2015)Panayotov, Chen, Povey, and Khudanpur]{panayotov2015librispeech}
Vassil Panayotov, Guoguo Chen, Daniel Povey, and Sanjeev Khudanpur.
\newblock Librispeech: an asr corpus based on public domain audio books.
\newblock In \emph{Acoustics, Speech and Signal Processing (ICASSP), 2015 IEEE International Conference on}, pages 5206--5210. IEEE, 2015.

\bibitem[Hartmann et~al.(2006)Hartmann, Bontas, Palma, and G{\'{o}}mez{-}P{\'{e}}rez]{HartmannBPG06}
Jens Hartmann, Elena~Paslaru Bontas, Ra{\'{u}}l Palma, and Asunci{\'{o}}n G{\'{o}}mez{-}P{\'{e}}rez.
\newblock {DEMO} - design environment for metadata ontologies.
\newblock In York Sure and John Domingue, editors, \emph{The Semantic Web: Research and Applications, 3rd European Semantic Web Conference, {ESWC} 2006, Budva, Montenegro, June 11-14, 2006, Proceedings}, volume 4011 of \emph{Lecture Notes in Computer Science}, pages 427--441. Springer, 2006.
\newblock \doi{10.1007/11762256\_32}.
\newblock URL \url{https://doi.org/10.1007/11762256\_32}.

\bibitem[Fleiss(1971)]{fleiss1971measuring}
Joseph~L Fleiss.
\newblock Measuring nominal scale agreement among many raters.
\newblock \emph{Psychological Bulletin}, 76\penalty0 (5):\penalty0 378--382, 1971.

\end{thebibliography}
\newpage%
\section*{Checklist}%

\begin{enumerate}

\item For all authors...
\begin{enumerate}
  \item Do the main claims made in the abstract and introduction accurately reflect the paper's contributions and scope?
    \answerYes{}
  \item Did you describe the limitations of your work?
    \answerYes{See \Cref{ssec:userstudy_results}}
  \item Did you discuss any potential negative societal impacts of your work?
    \answerNA{}
  \item Have you read the ethics review guidelines and ensured that your paper conforms to them?
    \answerYes{}
\end{enumerate}

\item If you are including theoretical results...
\begin{enumerate}
  \item Did you state the full set of assumptions of all theoretical results?
    \answerNA{}
	\item Did you include complete proofs of all theoretical results?
    \answerNA{}
\end{enumerate}

\item If you ran experiments (e.g. for benchmarks)...
\begin{enumerate}
  \item Did you include the code, data, and instructions needed to reproduce the main experimental results (either in the supplemental material or as a URL)?
    \answerNA{}
  \item Did you specify all the training details (e.g., data splits, hyperparameters, how they were chosen)?
    \answerNA{}
	\item Did you report error bars (e.g., with respect to the random seed after running experiments multiple times)?
    \answerNA{}
	\item Did you include the total amount of compute and the type of resources used (e.g., type of GPUs, internal cluster, or cloud provider)?
    \answerNA{}
\end{enumerate}

\item If you are using existing assets (e.g., code, data, models) or curating/releasing new assets...
\begin{enumerate}
  \item If your work uses existing assets, did you cite the creators?
    \answerNA{}
  \item Did you mention the license of the assets?
    \answerNA{}
  \item Did you include any new assets either in the supplemental material or as a URL?
    \answerYes{The Croissant format is accompanied by a technical specification, an editor, and an open-source library, referenced in \Cref{sec:intro} and \Cref{ssec:tools_integration}.}
  \item Did you discuss whether and how consent was obtained from people whose data you're using/curating?
    \answerYes{See \Cref{ssec:user_study}}
  \item Did you discuss whether the data you are using/curating contains personally identifiable information or offensive content?
    \answerNA{}
\end{enumerate}

\item If you used crowdsourcing or conducted research with human subjects...
\begin{enumerate}
  \item Did you include the full text of instructions given to participants and screenshots, if applicable?
    \answerYes{See Section \Cref{ssec:appendix_user_study} in supplemental materials.}
  \item Did you describe any potential participant risks, with links to Institutional Review Board (IRB) approvals, if applicable?
    \answerNA{}
  \item Did you include the estimated hourly wage paid to participants and the total amount spent on participant compensation?
    \answerNA{}
\end{enumerate}

\end{enumerate}

\begin{enumerate}

\item Submission introducing new datasets must include the following in the supplementary materials:
\begin{enumerate}
  \item Dataset documentation and intended uses. Recommended documentation frameworks include datasheets for datasets, dataset nutrition labels, data statements for NLP, and accountability frameworks.
  \item URL to website/platform where the dataset/benchmark can be viewed and downloaded by the reviewers.
  \item URL to Croissant metadata record documenting the dataset/benchmark available for viewing and downloading by the reviewers. You can create your Croissant metadata using e.g. the Python library available here: \href{https://github.com/mlcommons/croissant}{https://github.com/mlcommons/croissant}
  \item Author statement that they bear all responsibility in case of violation of rights, etc., and confirmation of the data license.
  \item Hosting, licensing, and maintenance plan. The choice of hosting platform is yours, as long as you ensure access to the data (possibly through a curated interface) and will provide the necessary maintenance.
\end{enumerate}

\item To ensure accessibility, the supplementary materials for datasets must include the following:
\begin{enumerate}
  \item Links to access the dataset and its metadata. This can be hidden upon submission if the dataset is not yet publicly available but must be added in the camera-ready version. In select cases, e.g when the data can only be released at a later date, this can be added afterward. Simulation environments should link to (open source) code repositories.
  \item The dataset itself should ideally use an open and widely used data format. Provide a detailed explanation on how the dataset can be read. For simulation environments, use existing frameworks or explain how they can be used.
  \item Long-term preservation: It must be clear that the dataset will be available for a long time, either by uploading to a data repository or by explaining how the authors themselves will ensure this.
  \item Explicit license: Authors must choose a license, ideally a CC license for datasets, or an open source license for code (e.g. RL environments).
  \item Add structured metadata to a dataset's meta-data page using Web standards (like schema.org and DCAT): This allows it to be discovered and organized by anyone. If you use an existing data repository, this is often done automatically.
  \item Highly recommended: a persistent dereferenceable identifier (e.g. a DOI minted by a data repository or a prefix on identifiers.org) for datasets, or a code repository (e.g. GitHub, GitLab,...) for code. If this is not possible or useful, please explain why.
\end{enumerate}

\item For benchmarks, the supplementary materials must ensure that all results are easily reproducible. Where possible, use a reproducibility framework such as the ML reproducibility checklist, or otherwise guarantee that all results can be easily reproduced, i.e. all necessary datasets, code, and evaluation procedures must be accessible and documented.

\item For papers introducing best practices in creating or curating datasets and benchmarks, the above supplementary materials are not required.
\end{enumerate}
\newpage%
\section{Appendix}%
\subsection{Code Examples}
\label{appendix0}

In this section, we illustrate through examples how users can work with Croissant files in their ML workflows, using the \texttt{mlcroissant} and TFDS libraries.

\subsubsection{Loading a Dataset from a Croissant File}
\label{appendix1}

\begin{mdframed}[backgroundcolor=black!5,leftmargin=0.5cm,skipabove=0.3cm,hidealllines=true,%
  innerleftmargin=0.1cm,innerrightmargin=0.2cm,innertopmargin=-0.0cm,innerbottommargin=-0.10cm]
\begin{lstlisting}[language=Python]
import mlcroissant as mlc
ds = mlc.Dataset("https://raw.githubusercontent.com/mlcommons/croissant/main/datasets/1.0/gpt-3/metadata.json")
metadata = ds.metadata.to_json()
print(f"{metadata['name']}:{metadata['description']}")
for x in ds.records(record_set="default"):
    print(x)
\end{lstlisting}
\end{mdframed}

\subsubsection{Loading data from a Croissant JSON-LD file in an ML workflow by using TFDS}
\label{appendix2}

\begin{mdframed}
[backgroundcolor=black!5,leftmargin=0.5cm,skipabove=0.3cm,hidealllines=true,%
  innerleftmargin=0.1cm,innerrightmargin=0.2cm,innertopmargin=-0.0cm,innerbottommargin=-0.10cm]
    \begin{lstlisting}[language=Python]
import tensorflow_datasets as tfds
builder = tfds.dataset_builders.CroissantBuilder(
    jsonld="https://raw.githubusercontent.com/mlcommons/croissant/main/datasets/0.8/huggingface-mnist/metadata.json",
    file_format="array_record",
)
builder.download_and_prepare()
ds = builder.as_data_source()
print(ds["default"][0])
\end{lstlisting}
\end{mdframed}

\subsubsection{Using Croissant into ML-workflow by loading into TFDS Data loader for HF Datasets}
\label{appendix3}

\begin{mdframed}
[backgroundcolor=black!5,leftmargin=0.5cm,skipabove=0.3cm,hidealllines=true,%
  innerleftmargin=0.1cm,innerrightmargin=0.2cm,innertopmargin=-0.0cm,innerbottommargin=-0.10cm]
\begin{lstlisting}[language=Python]
# 1. Point to a local or remote Croissant file
import mlcroissant as mlc
url = "https://huggingface.co/api/datasets/fashion_mnist/croissant"

# 2. Inspect metadata
print(mlc.Dataset(url).metadata.to_json())

# 3. Use Croissant dataset in your ML workload
import tensorflow_datasets as tfds
builder = tfds.core.dataset_builders.CroissantBuilder(
    jsonld=url,
    record_set_ids=["record_set_fashion_mnist"],
    file_format="array_record",
)
builder.download_and_prepare()

# 4. Split for training/testing
train, test = builder.as_data_source(
    split=["default[:80%
)
\end{lstlisting}
\end{mdframed}

\subsubsection{Visualizing Bounding Boxes in Croissant using the \href{https://cocodataset.org/}{COCO 2014 dataset}}
\label{appendix4}

\begin{mdframed}
[backgroundcolor=black!5,leftmargin=0.5cm,skipabove=0.3cm,hidealllines=true,%
  innerleftmargin=0.1cm,innerrightmargin=0.2cm,innertopmargin=-0.0cm,innerbottommargin=-0.10cm]
\begin{lstlisting}[language=Python]
# 1. Importing mlcroissant Python package
import mlcroissant as mlc

# 2. Create a subset of the COCO 2014 dataset which offers bounding box annotations
record_set = "images_with_bounding_box"

# We download resources from the validation split to download smaller files.
distribution = [
    mlc.FileObject(
        id="annotations_trainval2014.zip",
        name="annotations_trainval2014.zip",
        description="",
        content_url=(
        "http://images.cocodataset.org/annotations/annotations_trainval2014.zip",
    ),
    encoding_format="application/zip",
        sha256="031296bbc80c45a1d1f76bf9a90ead27e94e99ec629208449507a4917a3bf009",
    ),
    mlc.FileObject(
        id="annotations",
        name="annotations",
        description="",
        contained_in=["annotations_trainval2014.zip"],
        content_url="annotations/instances_val2014.json",
        encoding_format="application/json",
    ),
]
    
# The record set has the `image_id` and the `bbox` (short for bounding box).
record_sets = [
    mlc.RecordSet(
        id="images_with_bounding_box",
        name=record_set,
        fields=[
            mlc.Field(
                id="images_with_bounding_box/image_id",
                name="image_id",
                description="",
                data_types=mlc.DataType.INTEGER,
                source=mlc.Source(
                    file_object="annotations",
                    extract=mlc.Extract(
                        json_path="$.annotations[*].image_id"
                    ),
                ),
            ),
            mlc.Field(
                id="images_with_bounding_box/bbox",
                name="bbox",
                description="",
                data_types=mlc.DataType.BOUNDING_BOX,
                source=mlc.Source(
                    file_object="annotations",
                    extract=mlc.Extract(
                        json_path="$.annotations[*].bbox"
                    ),
                ),
            ),
        ],
    ),
]
    
metadata = mlc.Metadata(
    name="COCO2014",
    url="https://cocodataset.org",
    distribution=distribution,
    record_sets=record_sets,
)

# 3. Creating the Croissant JSON-LD file
jsonld = epath.Path("croissant.json")
with jsonld.open("w") as f:
    f.write(json.dumps(metadata.to_json(), indent=2))

# 4. Getting the first record from the generated Croissant JSON-LD
dataset = mlc.Dataset(jsonld=jsonld)
records = dataset.records(record_set=record_set)
record = next(iter(records))
print("The first record:")
print(json.dumps(record, indent=2))

# 5. Visualizing the bounding box
image_id, bbox = record["images_with_bounding_box/image_id"], record["images_with_bounding_box/bbox"]
url = f"http://images.cocodataset.org/val2014/COCO_val2014_{image_id:012d}.jpg"
    
# Download the image
print(f"Downloading {url}...")
response = requests.get(url)
image = Image.open(io.BytesIO(response.content))
draw = ImageDraw.Draw(image)
    
# COCO uses the XYWH format. PIL uses the XYXY format.
x1, y1, w, h = bbox
draw.rectangle((x1, y1, x1 + w, y1 + h), outline=(0, 255, 0), width=2)
display(image)
\end{lstlisting}
\end{mdframed}

\subsection{Croissant Health Metrics}
\label{appendix5}

Croissant Health is a framework to automatically scrape and compute metrics about Croissant from online dataset repositories. It has been implemented so far for Hugging Face Datasets and OpenML, and can be easily extended to new repositories. The metrics are derived from the crawl responses for hosted datasets and the number of \texttt{FileObject}s, \texttt{FileSet}s, \texttt{RecordSet}s, and \texttt{Field}s they contain. More detailed statistics will be added in the future.

\subsubsection{Croissant Statistics for Hugging Face Datasets}
\label{appendix6}
\begin{figure}[h!] 
\begin{minipage}{.49\linewidth}
    \centering
    \includegraphics[width=0.85\columnwidth]{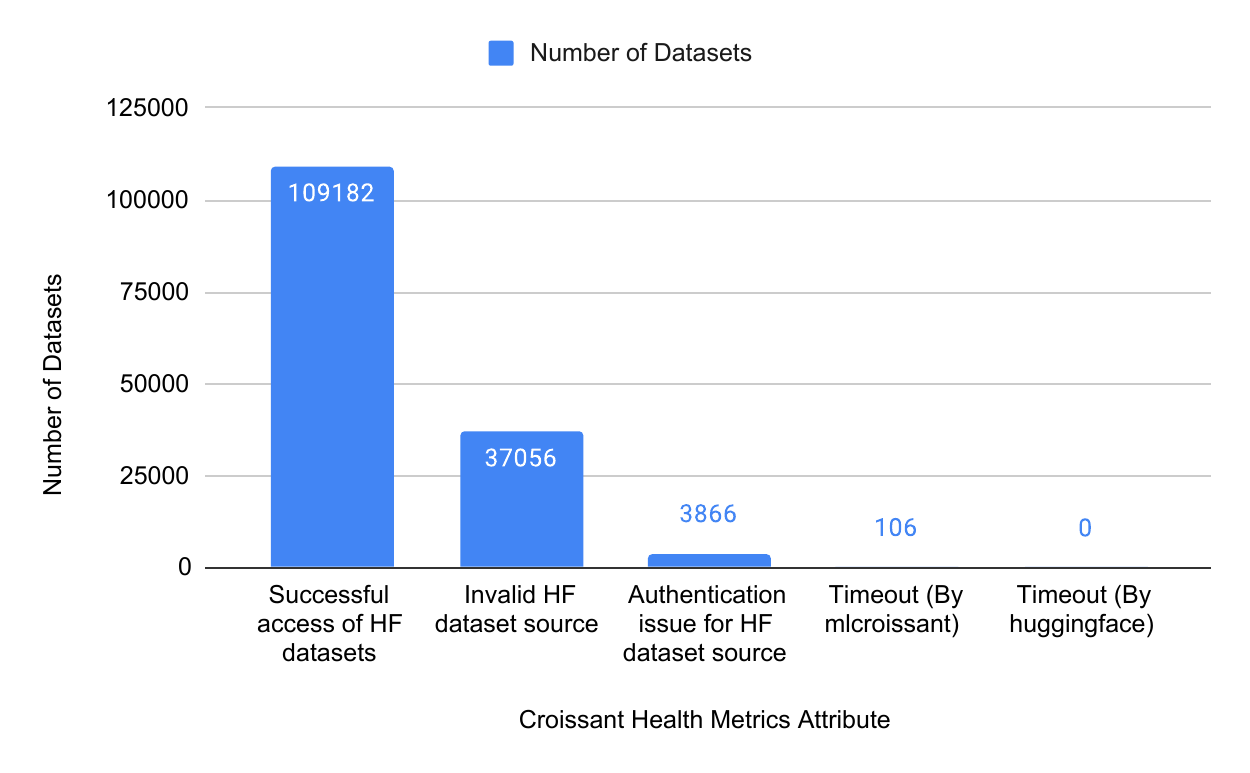}
    \caption{Scraping results for Croissant files of Hugging Face Datasets.}
    \label{fig:hf_scraping}
\end{minipage}%
\hspace{0.02\linewidth}
\begin{minipage}{.49\linewidth}
    \centering
    \includegraphics[width=0.95\columnwidth]{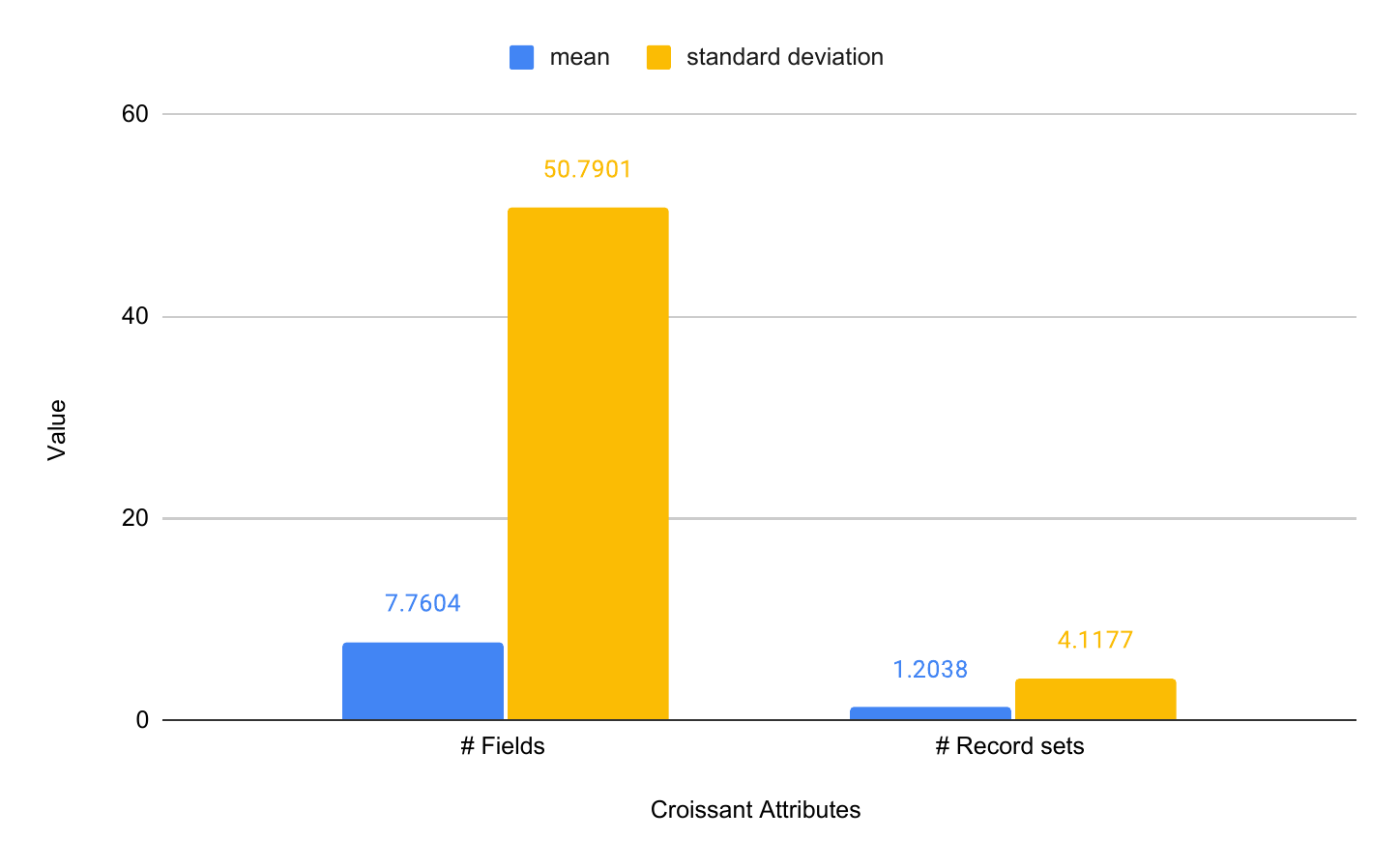}
    \caption{Illustration showing statistics for mean and standard deviation for the Croissant files hosted on Hugging Face datasets.}
    \label{fig:hf_chart}
\end{minipage}%
\end{figure}

Figure ~\ref{fig:hf_scraping} shows that the number of successfully downloaded Croissant datasets from Hugging Face is over 100k, and the  rate of invalid Croissant files is 25\%. These statistics are key to identify issues with Croissant generation and fix errors. Figure \ref{fig:hf_chart} gives an idea of the shape of these datasets: On average, datasets are small across all dimensions, with less than 10 resources, \texttt{RecordSet}s, and \texttt{Field}s.

\subsubsection{Croissant Stastistics for OpenML Datasets}
\label{appendix7}
\begin{figure}[h!] 
\begin{minipage}{.49\linewidth}
    \centering
    \includegraphics[width=0.85\columnwidth]{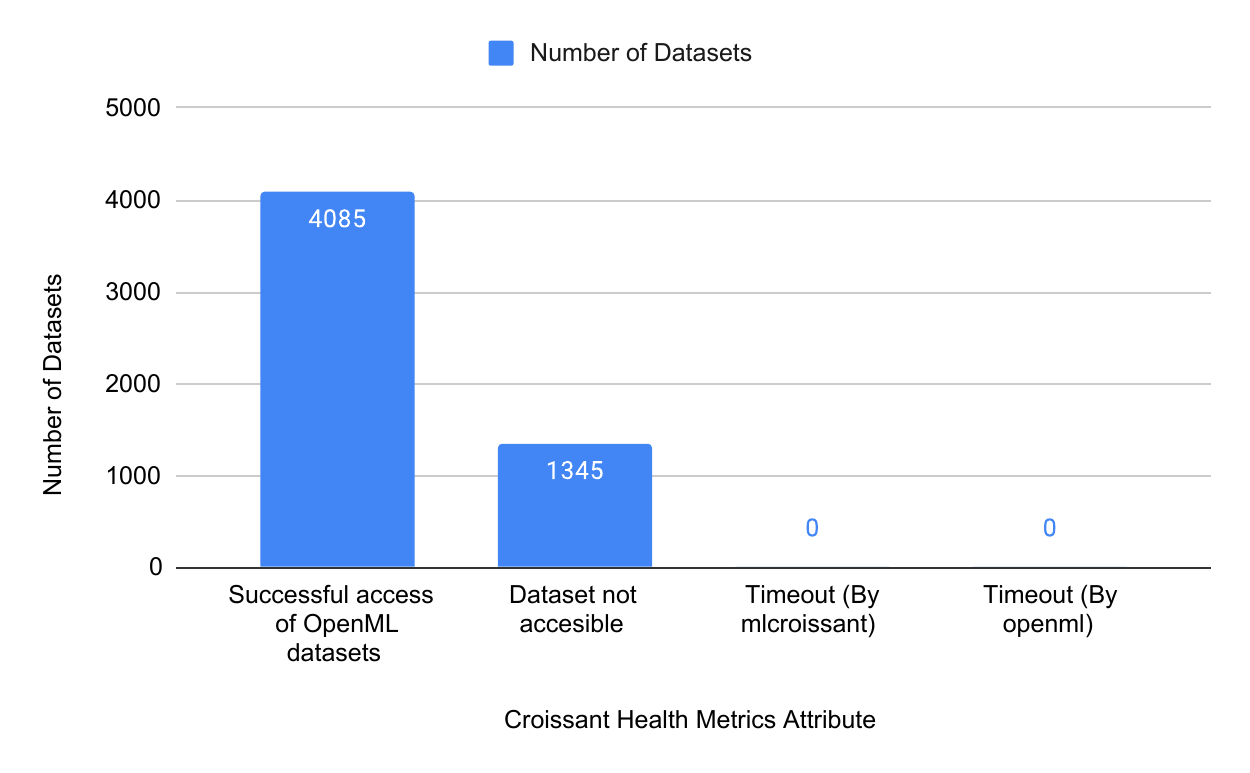}
    \caption{Scraping results for Croissant files hosted on OpenML.}
    \label{fig:openml_scraping}
\end{minipage}%
\hspace{0.02\linewidth}
\begin{minipage}{.49\linewidth}
    \centering
    \includegraphics[width=0.95\columnwidth]{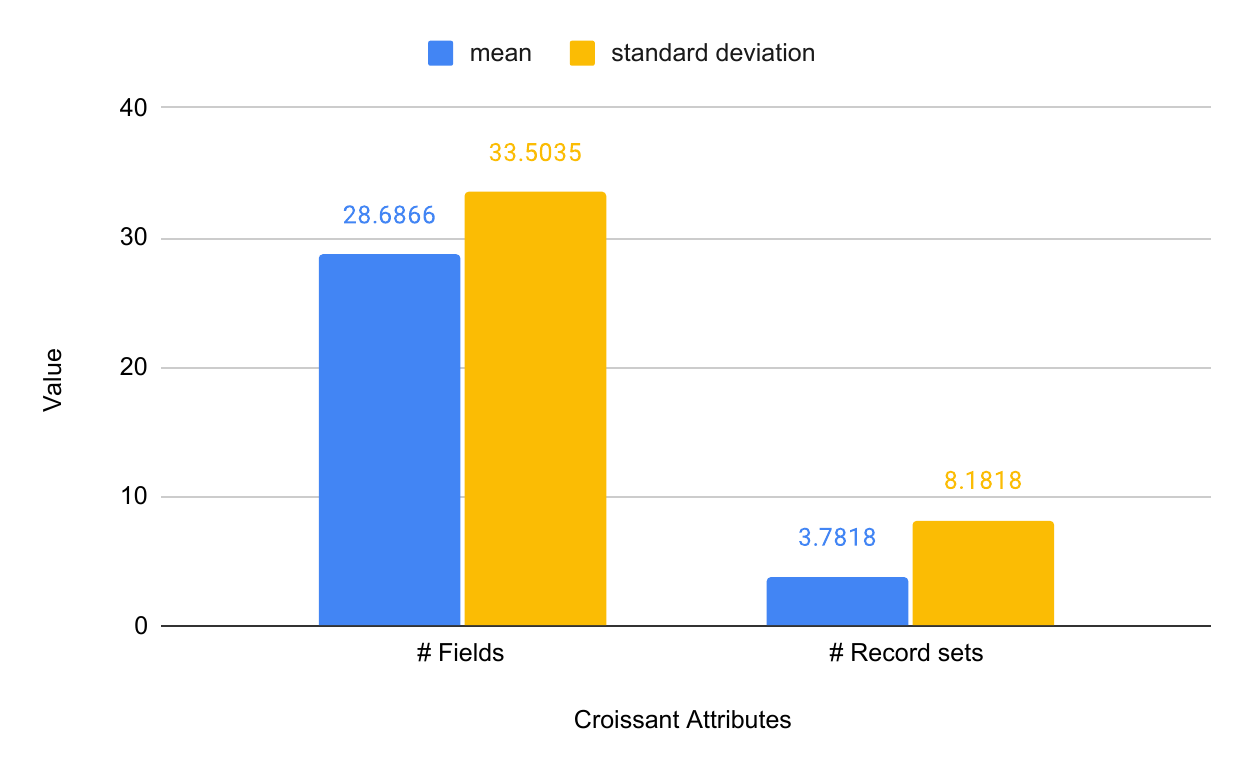}
    \caption{Illustration showing statistics for mean and standard deviation for Croissant files hosted on OpenML datasets.}
    \label{fig:openml_chart}
\end{minipage}%
\end{figure}

Figure \ref{fig:openml_scraping} shows the Croissant adoption for the OpenML datasets and Figure \ref{fig:openml_chart} illustrates the statistics for OpenML datasets. The number of datasets is much smaller overall, at about 4k datasets. The rate of invalid Croissant files is around 25\% due to authentication issues occurring while trying to access private datasets. We use Croissant Health\footnote{\url{https://github.com/mlcommons/croissant/tree/main/health}} to monitor the health of the Croissant ecosystem by crawling online JSON-LD files shared across repositories. Currently, Croissant Health performs this check for Hugging Face and OpenML datasets only, but will be extended in future to further repositories.\footnote{See \url{https://github.com/mlcommons/croissant/blob/main/health/visualizer/report_openml.ipynb} for further details.}

Figure \ref{fig:openml_chart} shows that these datasets are much more complex, with many datasets having a larger number or \texttt{FileObject}s, \texttt{RecordSet}s, and \texttt{Field}s compared to the Hugging Face ones.

\subsection{User Study}
\label{ssec:appendix_user_study}

\begin{figure}[t!] 
\begin{minipage}{.49\linewidth}
    \centering
    \includegraphics[width=1.0\columnwidth]{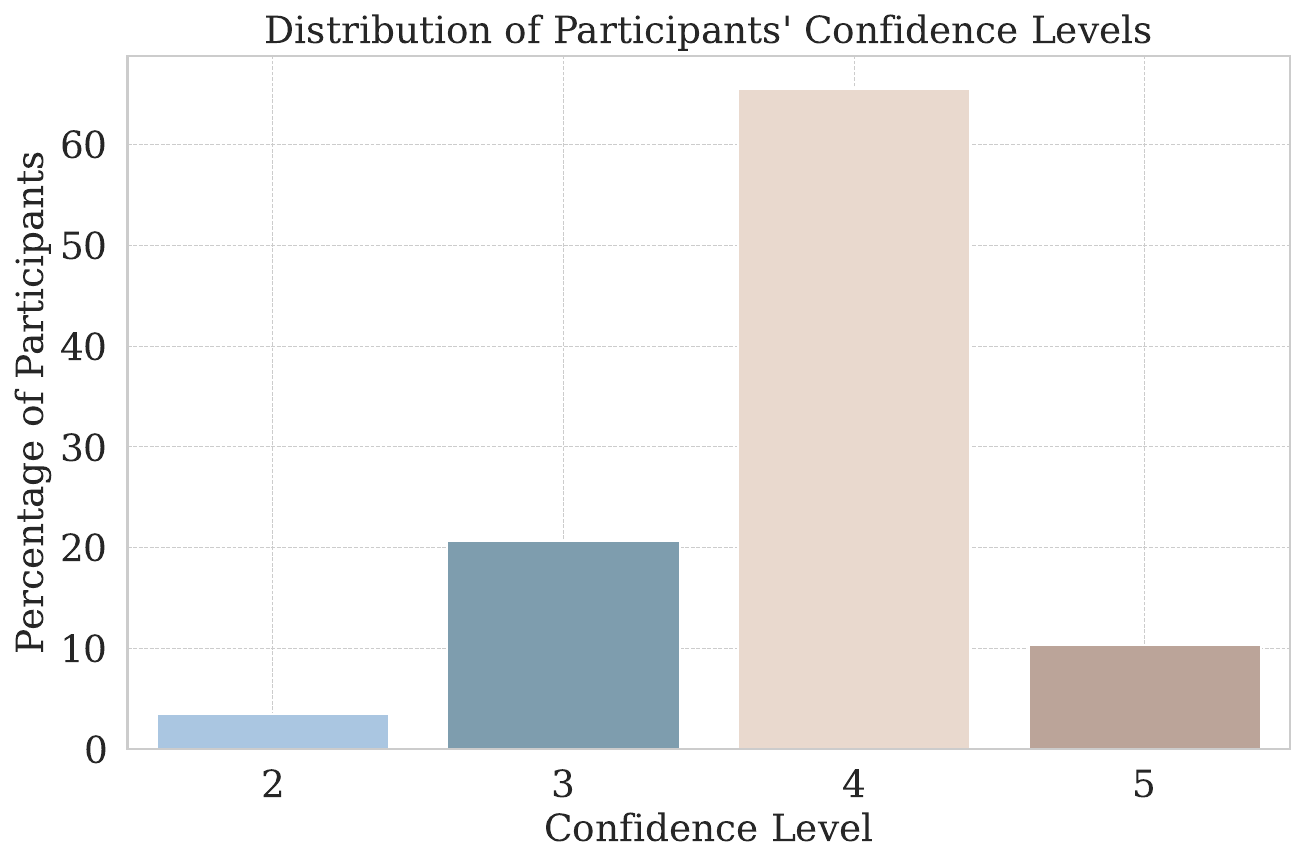}
    \caption{Annotators' confidence in provided annotations on a Likert scale from one to five. One indicates no confidence and five very high confidence in correct annotations.}
    \label{fig:confidence}
\end{minipage}%
\hspace{0.02\linewidth}
\begin{minipage}{.49\linewidth}
    \centering
    \includegraphics[width=1.0\columnwidth]{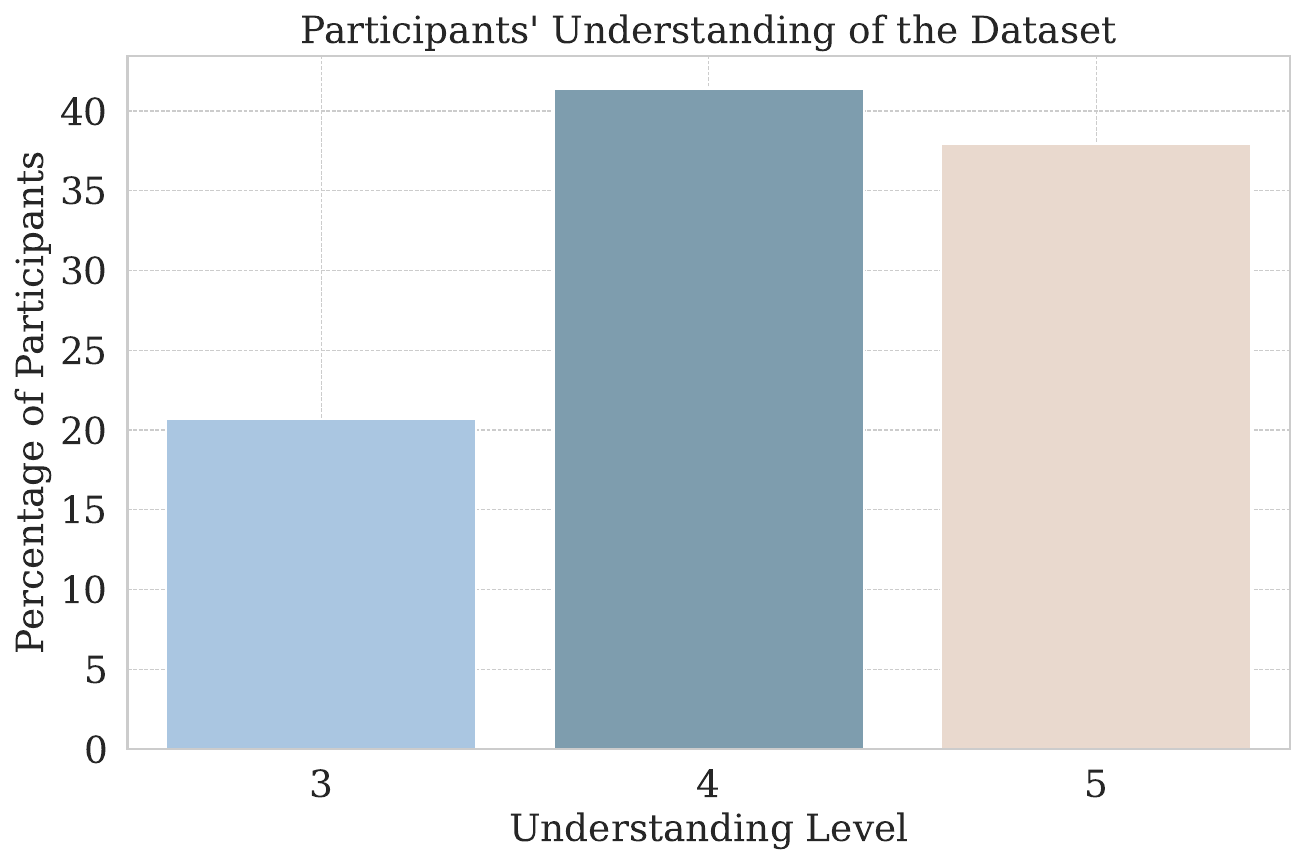}
    \caption{Annotators' understanding of datasets on a Likert scale from one to five. One indicates that the annotator has no understanding of the dataset while five means that the annotator understands the dataset, including its purpose, creation, etc.}
    \label{fig:understanding_dataset}
\end{minipage}%
\end{figure}

\Cref{fig:confidence} shows the participants' confidence in the annotations they provided, on a scale of 1 to 5. The majority of participants picked 4, which shows a high level of confidence in their ability to create Croissant metadata. It's interesting to contrast this number with the participants' level of understanding of datasets (\Cref{fig:understanding_dataset}), which varies more broadly between 3 and 5.

\begin{figure}[t!] 
    \centering
    \includegraphics[width=0.55\columnwidth]{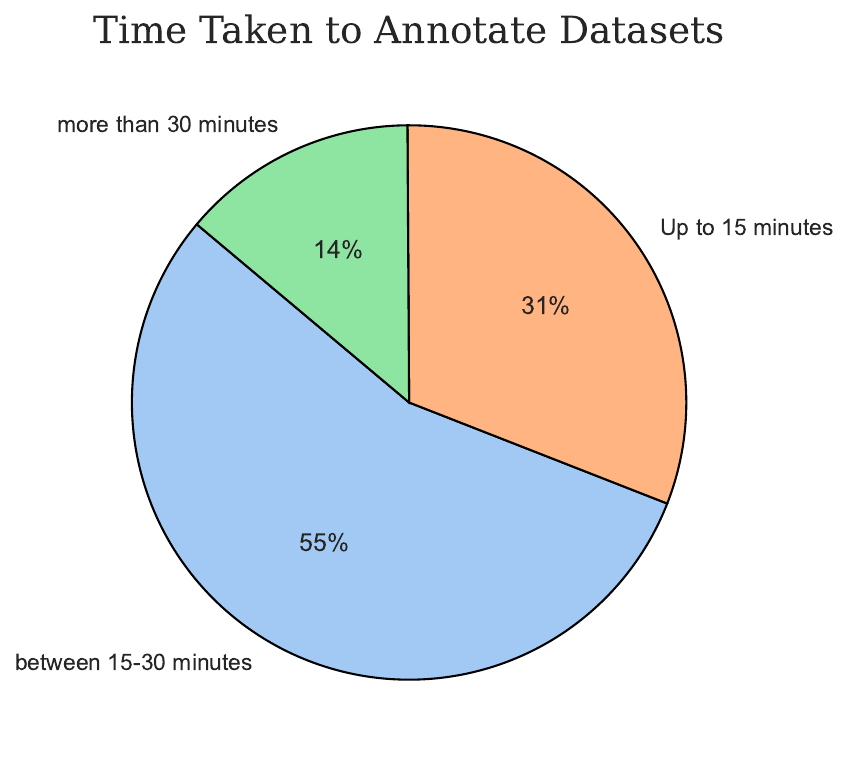}
    \caption{Time to create a Croissant description for a dataset.}
    \label{fig:time}
\end{figure}

\Cref{fig:time} gives an overview of how much time participants took for the user study. The majority of participants took 15-30 minutes to create the Croissant description of a dataset, which seems like a reasonable amount of time.

\begin{figure}[t!] 
    \centering
    \includegraphics[width=0.9\columnwidth]{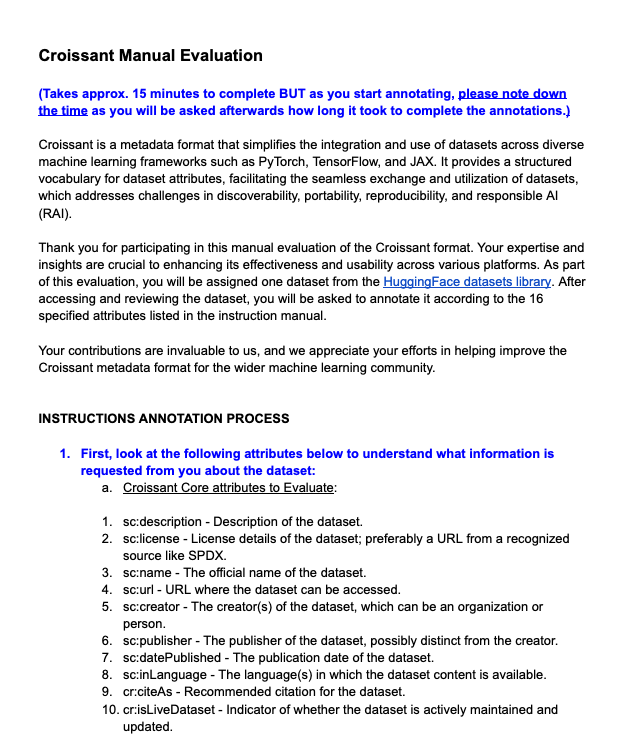}
    \caption{Instruction provided to user study participants for annotating ML datasets with selected Croissant/Croissant-RAI attributes (1/3).}
    \label{fig:userstudy_instruction_1}
\end{figure}

\begin{figure}[t!] 
    \centering
    \includegraphics[width=0.69\columnwidth]{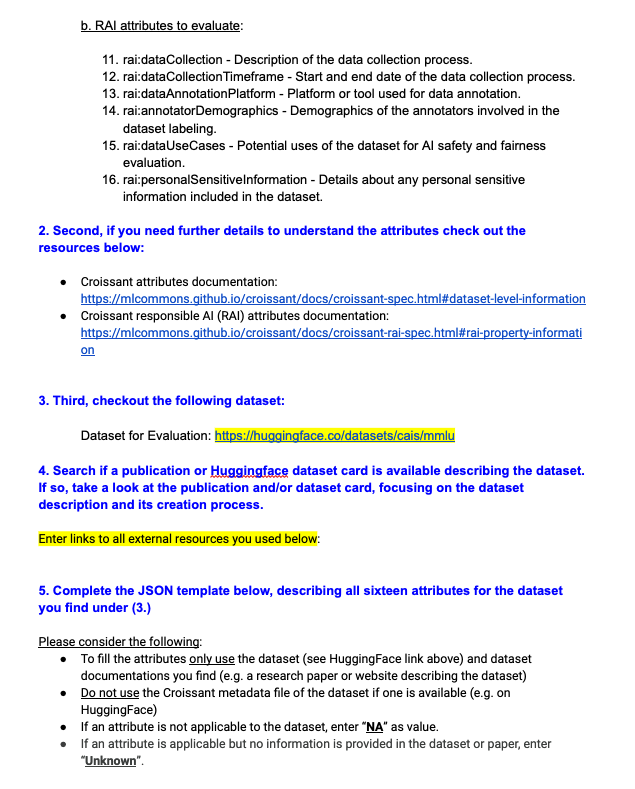}
    \caption{Instruction provided to user study participants for annotating ML datasets with selected Croissant/Croissant-RAI attributes (2/3).}
    \label{fig:userstudy_instruction_2}
\end{figure}

\begin{figure}[t!] 
    \centering
    \includegraphics[width=0.69\columnwidth]{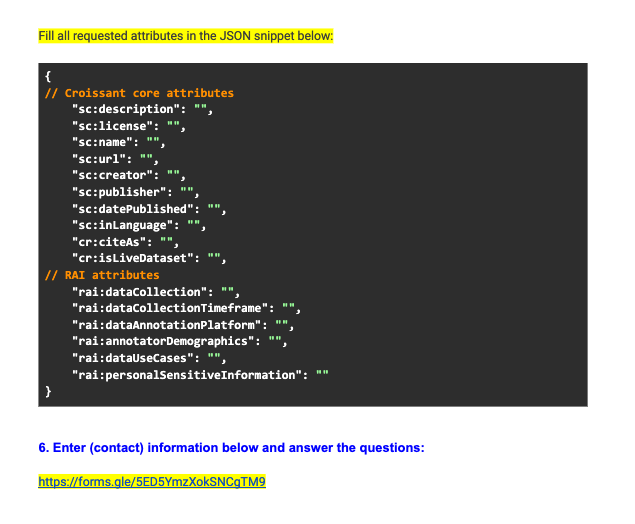}
    \caption{Instruction provided to user study participants for annotating ML datasets with selected Croissant/Croissant-RAI attributes (3/3).}
    \label{fig:userstudy_instruction_3}
\end{figure}

Finally, Figures \ref{fig:userstudy_instruction_1}, \ref{fig:userstudy_instruction_2}, and \ref{fig:userstudy_instruction_3} show the instruction provided to participants for the user study. 

\clearpage
\subsection{Semantic search with Croissant}
\label{appendix8}

The unified format of Croissant data makes it possible to scrape them from across the web and then conveniently embed and project them through a pipeline of your choice for semantic search among datasets. We provide a starter kit with an example of OpenML data at this address \url{https://github.com/mlcommons/croissant/tree/main/health/visualizer/explorer} where we
\begin{enumerate}
    \item Scrape Croissant files from the OpenML API following the steps under \url{https://github.com/mlcommons/croissant/tree/main/health}
    \item Read all Croissant dataset descriptions from the OpenML crawl (>5k)
    \item Extract dataset descriptions and urls from the Croissant files
    \item Project dataset descriptions onto an embedding space with a sentence transformer encoder
    \item Project embeddings to a three-dimensional space with PCA and t-SNE
    \item Explore semantic proximity of datasets in t-SNE embedding space
\end{enumerate}

An example visualizer can be found on \href{https://docs.mlcommons.org/croissant/}{https://docs.mlcommons.org/croissant/}.

\begin{figure}
    \centering
    \includegraphics[width=\linewidth]{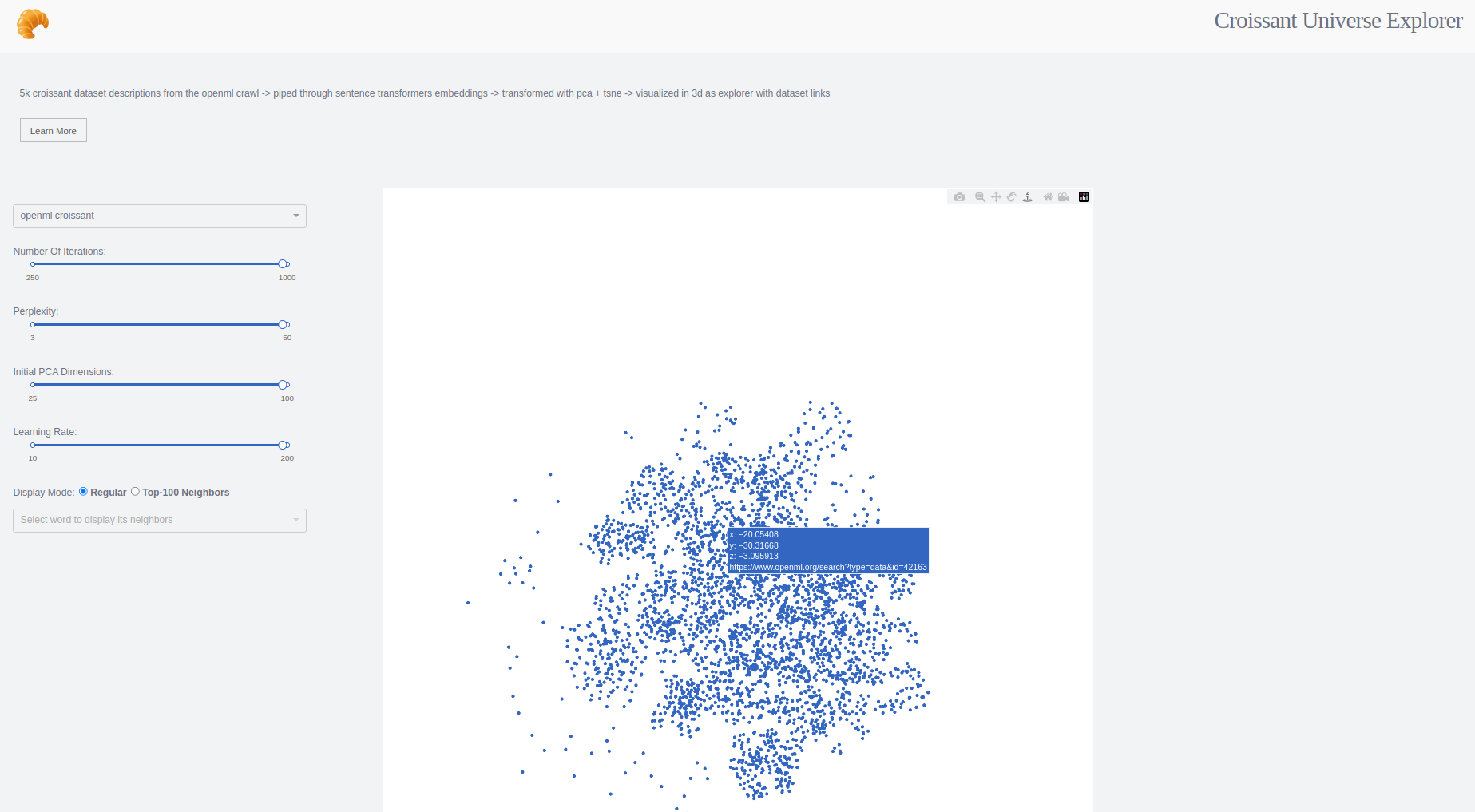}
    \caption{A visualizer example for exploring semantic similarity between datasets based on Croissant dataset Transformer and t-SNE embedding.}
    \label{fig:enter-label}
\end{figure}

\subsection{Croissant Editor for Dataset Authors}
\label{appendix9}
\begin{figure}
    \centering
    \includegraphics[width=0.9\linewidth]{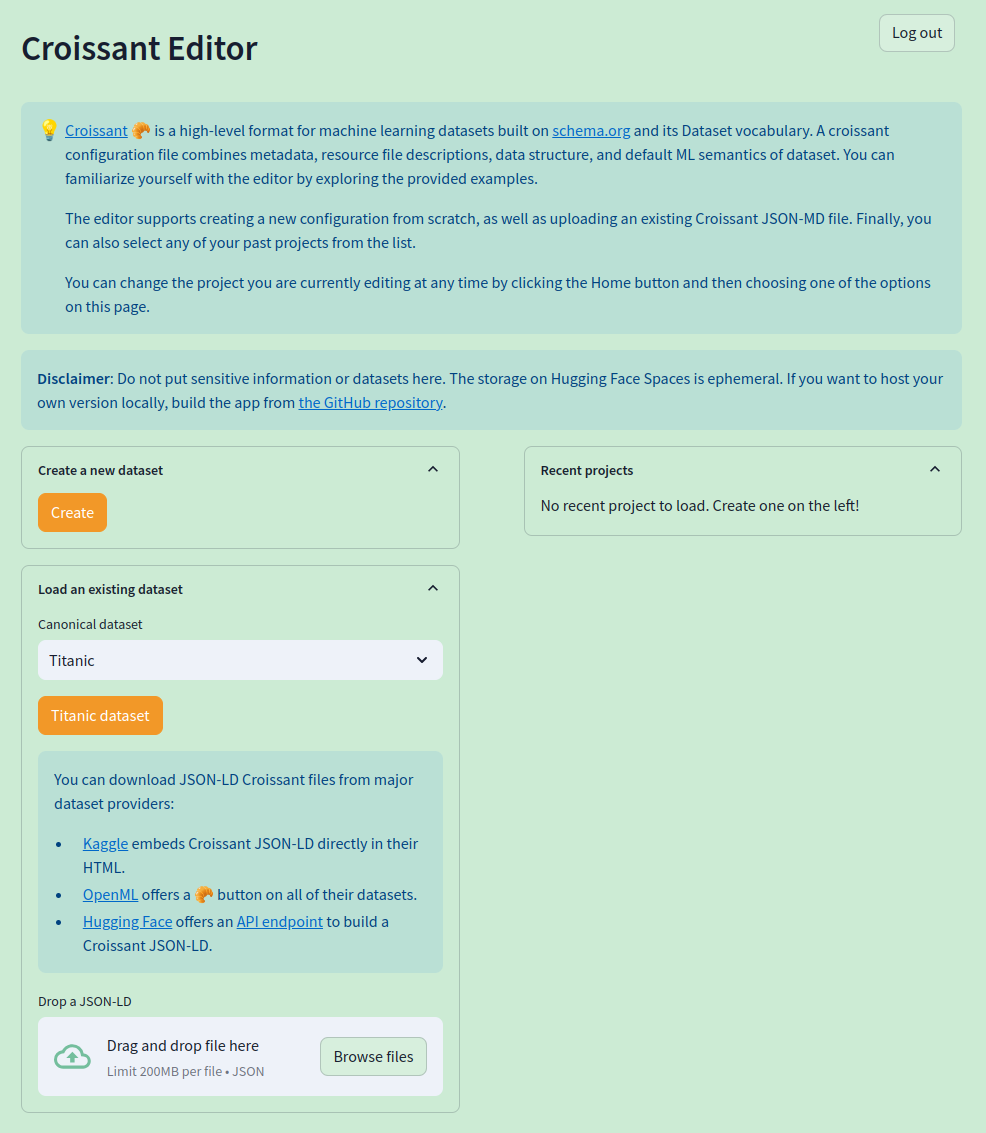}
    \caption{The Croissant editor Graphical User Interface (GUI).}
    \label{fig:editor-1}
\end{figure}

The Croissant open-source editor (Figure ~\ref{fig:editor-1}) is a tool for generating Croissant metadata for dataset publishers. The editor abstracts away
the details of the Croissant syntax via a familiar user interface. Users can drag-and-drop files to start creating a Croissant dataset. 

\begin{figure}
    \centering
    \includegraphics[width=0.9\linewidth]{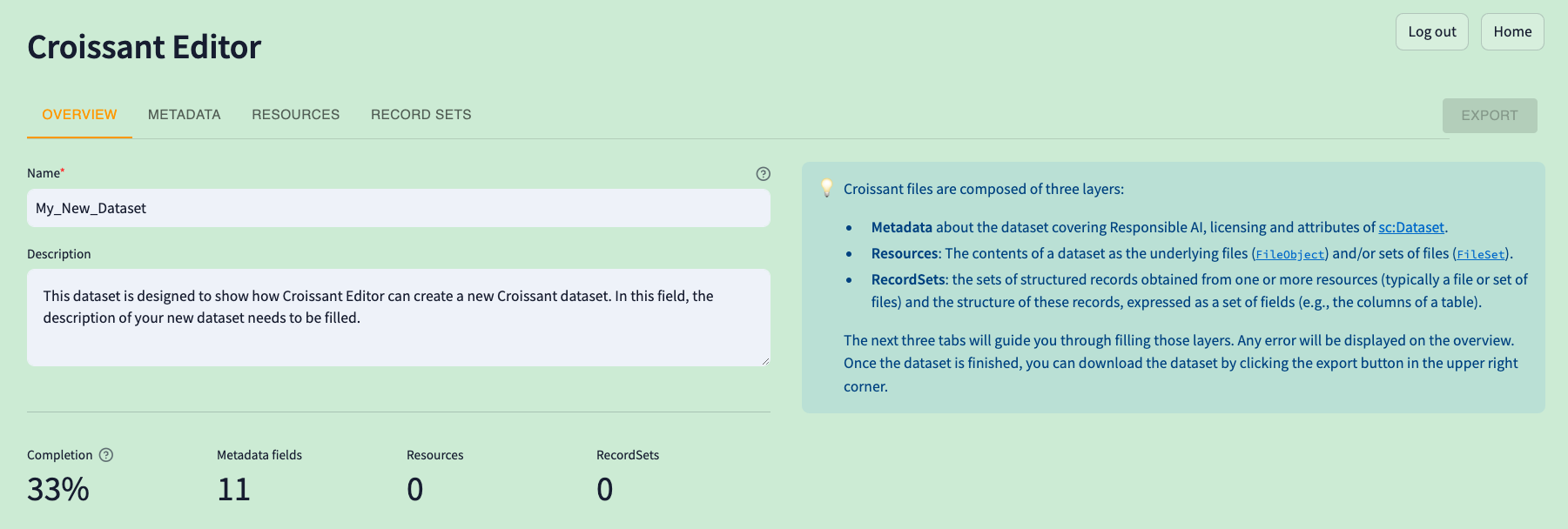}
    \includegraphics[width=0.9\linewidth]{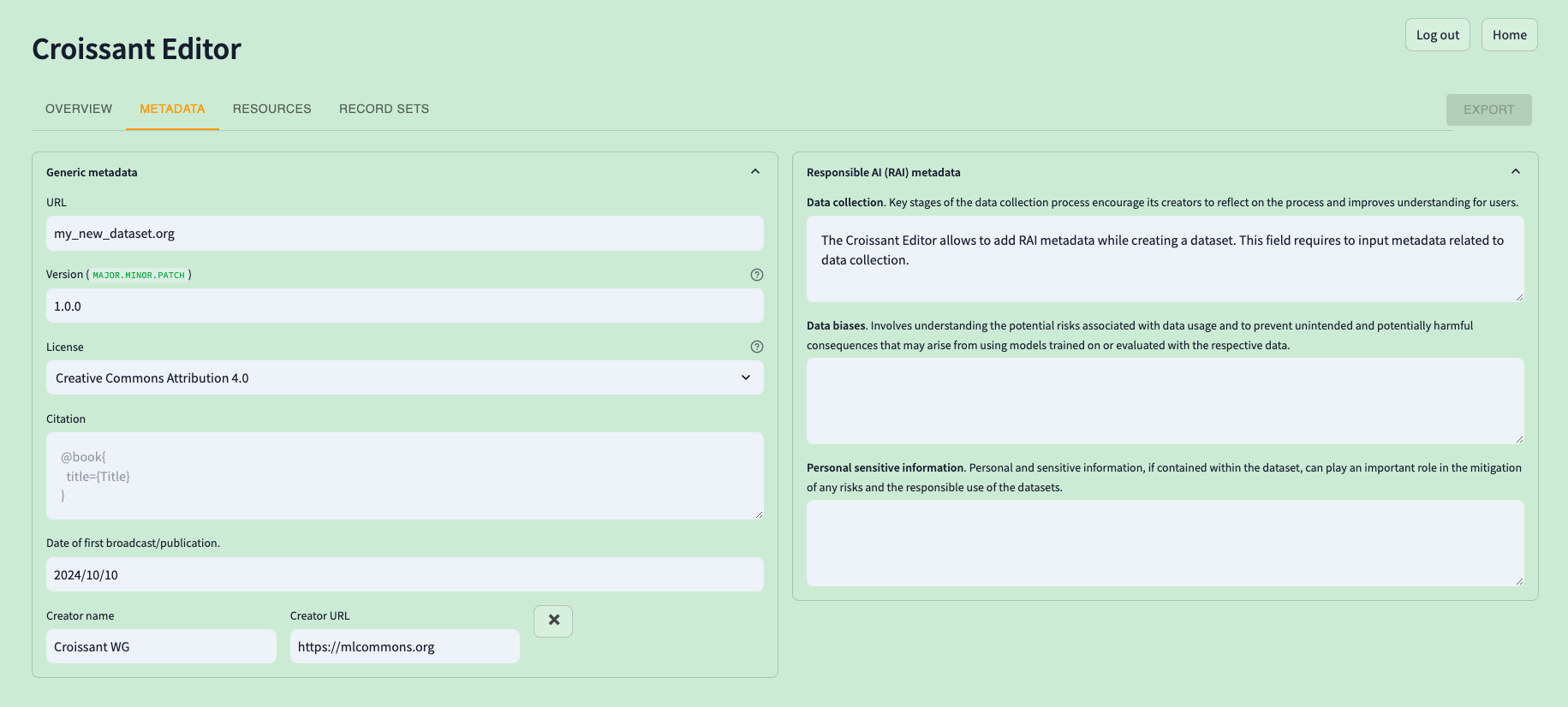}
    \caption{Illustration of the Croissant editor GUI for filling dataset and Responsible AI (RAI) attributes.}
    \label{fig:editor-2}
\end{figure}

The editor infers the resources and structure definitions from the data, and guides them in filling out required and optional fields (Figure ~\ref{fig:editor-2}). The editor can be run locally as well as on the Hugging Face interface and incorporates Croissant Core and Croissant RAI attributes (Figure ~\ref{fig:editor-2}) for generating Croissant file while hosting a dataset \footnote{\href{Croissant Editor}{\href{https://HuggingFace.co/spaces/MLCommons/croissant-editor}{https://HuggingFace.co/spaces/MLCommons/croissant-editor}}}.%
\end{document}